\documentclass{tut2016}

\usepackage[colorlinks=true,breaklinks=true,
			bookmarks=false,urlcolor=blue,
			citecolor=blue,linkcolor=blue,
			bookmarksopen=false,draft=false]{hyperref}

\usepackage{amsfonts,makecell}
\usepackage{authblk, dsfont, mathrsfs, bm, enumitem, nicefrac}
\usepackage{subfigure, transparent} 
\usepackage{algorithm, algorithmic, multirow, titlecaps, accents}
\usepackage[justification=centering]{caption}

\DeclareMathOperator{\diff}{d}
\DeclareMathOperator{\Lip}{Lip}

\DeclareMathOperator{\Trace}{Tr}
\DeclareMathOperator{\st}{\;s.t.}

\newcommand{\Sup}{\sup\limits}
\newcommand{\Inf}{\inf\limits}

\newcommand{\PP}{\mathds{P}}
\newcommand{\QQ}{\mathds{Q}}
\newcommand{\EE}{\mathds{E}}
\newcommand{\RR}{\mathbb{R}}
\newcommand{\opt}{^\star}

\newcommand{\Lc}{\mathcal{L}}

\newcommand{\risk}{\mathcal{R}}
\newcommand{\PSD}{\mathbb{S}_{+}}

\newcommand{\Pnom}{\widehat \PP}
\newcommand{\Tr}[1]{\Trace \left[ #1 \right]}

\newcommand{\optimize}[1]{\begin{array}{c@{\quad}l@{\quad~}l} #1 \end{array}}

\newcommand{\eps}{\varepsilon}
\newcommand{\cov}{\Sigma}
\newcommand{\covsa}{\widehat\cov}
\newcommand{\m}{\mu}
\newcommand{\msa}{\widehat\m}
\newcommand{\dualvar}{\gamma}
\newcommand{\B}{\mathbb{B}}

\newcommand{\eig}{\lambda}

\begin{document}

\CHAPTERNO{\phantom{Chapter 1}}
\DOI{}

\TITLE{Wasserstein Distributionally Robust Optimization:
    Theory and Applications in Machine Learning}

\AUBLOCK{

\AUTHOR{Daniel Kuhn}
\AFF{Risk Analytics and Optimization Chair, EPFL, Lausanne 1015, Switzerland,
\EMAIL{daniel.kuhn@epfl.ch}}

\AUTHOR{Peyman Mohajerin Esfahani}
\AFF{Delft University of Technology, Mekelweg 2, 2628 CD, Delft, The Netherlands, \EMAIL{P.MohajerinEsfahani@tudelft.nl}}

\AUTHOR{Viet Anh Nguyen, Soroosh Shafieezadeh-Abadeh}
\AFF{Risk Analytics and Optimization Chair, EPFL, Lausanne 1015, Switzerland,
\textbraceleft\EMAIL{viet-anh.nguyen@epfl.ch}, \EMAIL{soroosh.shafiee@epfl.ch}\textbraceright}
}

\CHAPTERHEAD{Wasserstein Distributionally Robust Optimization}

\ABSTRACT{Many decision problems in science, engineering and economics are affected by uncertain parameters whose distribution is only indirectly observable through samples. The goal of data-driven decision-making is to learn a decision from finitely many training samples that will perform well on unseen test samples. This learning task is difficult even if all training and test samples are drawn from the same distribution---especially if the dimension of the uncertainty is large relative to the training sample size. Wasserstein distributionally robust optimization seeks data-driven decisions that perform well under the most adverse distribution within a certain Wasserstein distance from a nominal distribution constructed from the training samples. In this tutorial we will argue that this approach has many conceptual and computational benefits. Most prominently, the optimal decisions can often be computed by solving tractable convex optimization problems, and they enjoy rigorous out-of-sample and asymptotic consistency guarantees. We will also show that Wasserstein distributionally robust optimization has interesting ramifications for statistical learning and motivates new approaches for fundamental learning tasks such as classification, regression, maximum likelihood estimation or minimum mean square error estimation, among others.

}

\maketitle

\vspace{-1.5em}
\begin{table}[h]
    \centering
    \begin{tabular}{ll}
        {\bf Update} &  Several errors in the published version \cite{kuhn2019wasserstein} of this paper have been corrected in\! \\
        & this version. For transparency, all updates are highlighted with a footnote. \\\Xhline{1pt}
    \end{tabular}
\end{table}

\section{Introduction}
\label{sec:introduction}
We consider a decision problem under uncertainty, where each admissible decision results in an uncertain loss that is modeled by a measurable extended real-valued {\em loss function} $\ell(\xi)$. We assume that the random vector $\xi\in\RR^m$ captures all decision-relevant risk factors and is governed by a probability distribution $\PP$. The feasible set of all available loss functions is denoted by $\Lc$. The {\em risk} of a decision $\ell\in\Lc$ is defined as the expected loss under $\PP$, that is,
\begin{equation}
\label{eq:risk}
\risk (\PP, \ell) = \EE^{\PP} [\ell(\xi)],
\end{equation}
and the {\em optimal risk} is defined as the risk of the least risky admissible loss function, that is,
\begin{equation}
\label{eq:opt:risk}
\risk(\PP, \Lc) = \inf_{\ell \in \Lc} ~ \risk (\PP, \ell).
\end{equation}
To ensure that the expectations in~\eqref{eq:risk} and~\eqref{eq:opt:risk} are defined for all measurable loss functions, we set $\EE^{\PP} [\ell(\xi)]=\infty$ whenever the expectations of the positive and negative parts of $\ell(\xi)$ are both infinite. This convention means that infeasibility trumps unboundedness.

In most real decision-making situations, the distribution $\PP$ is fundamentally unknown. However, $\PP$ may be indirectly observable through {\em training samples} $\widehat \xi_i$, $i\in\{1, \ldots, N\}$, drawn independently from $\PP$. In addition, some structural properties of $\PP$ may be known. For example, if $\xi$ represents a vector of uncertain prices, then $\PP$ must be supported on the nonnegative orthant~$\RR^m_+$. Alternatively, $\PP$ may be known to display certain symmetry or unimodality properties, or it may even be known to belong to some parametric distribution~family. 

If the distribution $\PP$ is unknown, we lack an important input parameter for the risk evaluation problem~\eqref{eq:risk} and the decision problem~\eqref{eq:opt:risk}. In this case, the unknown true distribution~$\PP$ could be replaced with a {\em nominal distribution} $\Pnom_N$ estimated from the $N$ training samples. Note that unlike $\PP$, the nominal distribution $\Pnom_N$ is accessible as it is constructed from observable quantities. Therefore, the nominal risk evaluation and decision problems (that is, problems~\eqref{eq:risk} and~\eqref{eq:opt:risk} with $\Pnom_N$ instead of $\PP$) are at least in principle solvable. The following example showcases common methods for constructing the nominal distribution~$\Pnom_N$.


\begin{example}[Nominal distribution]
    \label{ex:nominal-distribution}
    In the remainder we will primarily work with the following {\em non-parametric} and {\em parametric} models for the nominal distribution.
    \begin{enumerate}
        \item In the absence of any structural information, it is convenient to set $\Pnom_N$ to the discrete {\em empirical distribution}, that is, the uniform distribution on the $N$ training samples,
        \begin{equation}
        \label{eq:empirical}
        \Pnom_N = \frac{1}{N} \sum_{i=1}^N \delta_{\widehat \xi_i},
        \end{equation}
        where $\delta_{\widehat \xi_i}$ denotes the Dirac point mass at the $i^{\rm th}$ training sample $\widehat \xi_i$.
        \item 
            We say that $\QQ=\mathcal{E}_g(\mu, \cov)$ is an {\em elliptical} probability distribution if it has a density function of the form $f(\xi) = C \det(\cov)^{-1} \, g((\xi-\mu) \cov^{-1} (\xi-\mu))$ with density generator $g(u)\ge 0$ for all $u\ge 0$, normalization constant $C>0$, mean vector $\mu\in\RR^m$ and covariance matrix $\cov\in\mathbb{S}_{++}^m$. Examples of elliptical distributions are reported in Table~\ref{tabel:elliptical} of Appendix~\ref{sect:elliptical}. In the presence of structural information, it is often convenient to set $\Pnom_N$ to an elliptical distribution with a structure-dependent density generator $g$, that is,
        \begin{equation}
        \label{eq:elliptical}
        \Pnom_N = \mathcal{E}_g(\msa, \covsa),
        \end{equation}
         where only the mean vector $\msa$ and the covariance matrix $\covsa$ depend on the training samples and are constructed via maximum likelihood estimation. 
    \end{enumerate}
    As a function of the training data, the nominal distribution $\Pnom_N$ constitutes itself a random object, which is governed by the distribution $\PP^N$ of the $N$ independent training samples. \hfill $\square$
\end{example}


    
Even if the most sophisticated statistical tools are deployed, the nominal distribution $\Pnom_N$ will invariably differ from the unknown true distribution $\PP$ that generated the training samples. Moreover, if $\Pnom_N$ is used instead of $\PP$, the solutions of the risk evaluation problem~\eqref{eq:risk} and the decision problem~\eqref{eq:opt:risk} are likely to inherit any estimation errors in~$\Pnom_N$. In the context of financial portfolio theory it has even been observed that estimation errors in the input parameters of an optimization problem are often amplified by the optimization  \cite{chopra2011errors, michaud1989enigma}. To make things worse, one can generally show that even if the distributional input parameters of a decision problem are unbiased, the optimization results tend to be optimistically biased. Thus, implementing the optimal decisions leads to disappointment in out-of-sample tests. In decision analysis this phenomenon is sometimes termed the {\em optimizer's curse} \cite{smith2006curse}, and in stochastic optimization it is referred to as the {\em optimization bias} \cite{homemdemello2014, shapiro2003}.

\begin{example}[Optimizer's curse]
\label{ex:optimizers-curse}
Let $\Pnom_N$ be an unbiased estimator for $\PP$. Thus, we have $\EE^{\PP^N}[\Pnom_N]=\PP$, where the expectation is taken with respect to the distribution $\PP^N$ of the $N$ independent training samples. Then, the risk of a fixed loss function $\ell\in\Lc$ satisfies
\[
    \EE^{\PP^N}\left[\risk(\Pnom_N,\ell) \right] = \EE^{\PP^N}\left[ \EE^{\Pnom_N} [\ell(\xi)]\right] = \EE^{\PP} [\ell(\xi)] = \risk(\PP,\ell),
\]
where the second equality holds because the inner expectation is linear in $\Pnom_N$. This implies that $\risk(\Pnom_N,\ell)$ constitutes an unbiased estimator for the true risk $\risk(\PP,\ell)$. Moreover, we have
\[
    \EE^{\PP^N}\left[\risk(\Pnom_N,\Lc) \right] = \EE^{\PP^N}\left[ \inf_{\ell \in \Lc} ~ \risk (\Pnom_N, \ell) \right] \le \inf_{\ell \in \Lc} ~\EE^{\PP^N}\left[  \risk (\Pnom_N, \ell) \right] 
    =  \inf_{\ell \in \Lc} ~ \risk (\PP, \ell) = \risk(\PP,\Lc),
\]
where the inequality holds because the infimum inside the expectation can adapt to $\Pnom_N$. Hence, $\risk(\Pnom_N,\Lc)$ constitutes an optimistically biased estimator for $\risk(\PP,\Lc)$, {\em i.e.}, it underestimates the true risk. On the other hand, any optimizer $\ell^\star\in\arg\min_{\ell\in\Lc} \risk(\Pnom_N,\ell)$ satisfies
\[\
    \risk(\PP,\ell^\star)\ge \inf_{\ell\in\Lc} ~\risk(\PP,\ell) = \risk(\PP, \Lc).
\]
The above observations can be interpreted as follows. Someone solving the nominal decision problem {\em thinks} that the risk of $\ell^\star$ amounts to $\risk(\Pnom_N,\ell^\star) = \risk(\Pnom_N,\Lc)$ (the in-sample risk), which is typically {\em smaller} than the optimal risk $\risk(\PP,\Lc)$ attainable under full knowledge of~$\PP$. However, the {\em actual} risk $\risk(\PP,\ell^\star)$ of the optimizer~$\ell^\star$ under the true distribution (the out-of-sample risk) is always {\em larger} than $\risk(\PP,\Lc)$. The difference between the out-of-sample risk and the in-sample risk is termed the {\em post-decision disappointment}. The {\em optimizer's curse} refers to the observation that the post-decision disappointment is positive on average.
\hfill $\square$
\end{example}

In order to quantify the sensitivity of $\risk (\PP, \ell)$ and $\risk(\PP, \Lc)$ with respect to the unknown true distribution $\PP$, we must introduce a distance measure between probability distributions. As we will argue below, the Wasserstein distance is a particularly convenient choice.

\begin{definition}[Wasserstein distance]
    \label{definition:wasserstein}
    For any $p \in [1, \infty)$, the type-$p$ Wasserstein distance between two probability distributions $\QQ$ and $\QQ'$ on $\RR^m$ is defined as
    \begin{equation}
    \label{eq:wasserstein}
    W_p(\QQ, \QQ') = \left( \inf_{\pi \in \Pi(\QQ, \QQ')} ~ \int_{\RR^m \times \RR^m} \| \xi - \xi' \|^p \, \pi (\diff \xi, \diff \xi') \right)^{\frac{1}{p}},
    \end{equation}
    where $\| \cdot \|$ is a norm on $\RR^m$, while $\Pi(\QQ, \QQ')$ denotes the set of all joint probability distributions of $\xi\in\RR^m$ and $\xi'\in\RR^m$ with marginals $\QQ$ and $\QQ'$, respectively.
\end{definition}

One can show that the Wasserstein distance is a metric, that is, it is nonnegative, symmetric and subadditive, and it vanishes only if $\QQ=\QQ'$ \cite[p.~94]{villani2008optimal}. One can further show that $W_p(\QQ, \QQ')$ is finite whenever both $\QQ$ and $\QQ'$ have finite $p^{\rm th}$-order moments \cite[p.~95]{villani2008optimal}.

The optimal value of the optimization problem in~\eqref{eq:wasserstein} can be interpreted as the minimum cost of turning one pile of dirt represented by~$\QQ$ into another pile of dirt represented by~$\QQ'$, where the cost of moving a unit mass from $\xi$ to $\xi'$ amounts to $\|\xi-\xi'\|^p$. The decision variable $\pi$ thus encodes a transportation plan, that is, for any measurable sets $A,B\subseteq \RR^m$ the probability $\pi(A\times B)$ reflects the amount of mass that is moved from the source region~$A$ to the target region~$B$. Because of this interpretation, the Wasserstein distance is often referred to as the earth mover's distance in statistics and computer science~\cite{rubner2000earth}. The theory of optimal transport was pioneered by Monge in 1781 \cite{monge1781memoire} and formalized by Kantorovich in 1942 \cite{kantorovich1942translocation}. Accordingly, the Wasserstein distance is often referred to as the Monge-Kantorovich distance~\cite{villani2008optimal}. The Wasserstein distance is used in many areas of science.  In the wider context of machine learning, for instance, the Wasserstein distance is used for the analysis of mixture models~\cite{kolouri2018sliced, nguyen2013convergence} as well as for image processing~\cite{alvarez2018structured, ferradans2014regularized, kolouri2015transport, papadakis2017convex, tartavel2016wasserstein}, computer vision and graphics~\cite{pele2008linear, pele2009fast, rubner2000earth, solomon2015convolutional, solomon2014earth}, data-driven bioengineering~\cite{feydy2017optimal, kundu2018discovery, wang2011optimal}, clustering~\cite{ho2017multilevel}, dimensionality reduction~\cite{cazelles2018geodesic, flamary2018wasserstein, rolet2016fast, schmitz2018wasserstein, seguy2015principal}, deep learning with generative adversarial networks~\cite{arjovsky2017wasserstein, genevay2018learning, gulrajani2017improved}, domain adaptation~\cite{courty2017optimal, murez2018image}, signal processing~\cite{thorpe2017transportation}, etc. For a comprehensive survey of different applications of the optimal transport problem see~\cite{kolouri2017optimal, peyre2018computational}.

The optimization problem in~\eqref{eq:wasserstein} constitutes an infinite-dimensional linear program over the transportation plan~$\pi$. This linear program admits a strong dual, which in turn provides an alternative characterization of the Wasserstein distance.

\begin{theorem}[Dual Kantorovich problem]
    \label{thm:dual-kantorovich}
    For any $p \in [1, \infty)$, the $p^{\rm th}$ power of the type-$p$ Wasserstein distance between $\QQ$ and $\QQ'$ admits the dual representation
    \begin{equation*}
    \optimize{
        \displaystyle W_p^p(\QQ, \QQ') =~ \sup & \displaystyle \int_{\RR^m} \psi(\xi') \, \QQ'(\diff \xi') - \int_{\RR^m} \phi(\xi) \, \QQ(\diff \xi)  \\ [1em]
        \hspace{6.0em}\st & \phi \text{ and }\psi \text{ are bounded continuous functions on $\RR^m$ with} \\ [0.5em]
        & \displaystyle \psi(\xi)-\phi(\xi') \leq \| \xi - \xi' \|^p \quad \forall \xi, \xi' \in \RR^m.
    }
    \end{equation*}
\end{theorem}

For a proof of Theorem~\ref{thm:dual-kantorovich} see~\cite[\S~5]{villani2008optimal}. The dual problem can be interpreted as the profit maximization problem of a third party that reallocates the dirt from $\QQ$ to $\QQ'$ on behalf of the problem owner by buying dirt at the origin $\xi$ at unit price $\phi(\xi)$ and selling dirt at the destination $\xi'$ at unit price $\psi(\xi')$. The constraints ensure that the problem owner prefers to use the services of the third party for every origin-destination pair $(\xi,\xi')$ instead of reallocating the dirt independently at her own transportation cost $\|\xi-\xi'\|^p$. The optimal price functions $\phi^\star$ and $\psi^\star$---if they exist---are called Kantorovich potentials \cite[p.~99]{villani2008optimal}.

If $p=1$, the dual problem can be further simplified. To see this, we define the Lipschitz modulus of an extended real-valued function $\phi$ on $\RR^m$ with respect to the norm $ \| \cdot \|$ as 
$$\Lip(\phi) = \sup_{\xi \neq \xi'} ~ \frac{|\phi(\xi)-\phi(\xi')|}{\| \xi - \xi'\|}. $$ The Lipschitz modulus can be viewed as the slope of the steepest line segment connecting any two points on the graph of $\phi$. The following result simplifies Theorem~\ref{thm:dual-kantorovich} for $p=1$. 

\begin{theorem}[Kantorovich-Rubinstein theorem]
    \label{thm:kantorovich-rubinstein}
    The type-1 Wasserstein distance between $\QQ$ and $\QQ'$ admits the dual representation
    \[
    W_1(\QQ,\QQ') = \sup_{\Lip(\phi) \leq 1} ~ \int_{\RR^m} \phi(\xi) \, \QQ(\diff \xi) - \int_{\RR^m} \phi(\xi') \, \QQ'(\diff \xi').
    \]        
\end{theorem}

Kantorovich and Rubinstein~\cite{kantorovich1958space} originally established this result for compactly supported distributions. A modern proof for arbitrary distributions can be found in~\cite[Remark~6.5]{villani2008optimal}. Theorem~\ref{thm:kantorovich-rubinstein} asserts that the type-1 Wasserstein distance between $\QQ$ and $\QQ'$ equals the difference between the expected values of a test function $\phi$ under $\QQ$ and $\QQ'$, respectively, maximized across all Lipschitz-continuous test functions with Lipschitz modulus of at most~1.


The Kantorovich-Rubinstein theorem enables us to estimate the sensitivity of $\risk (\PP, \ell)$ and $\risk(\PP, \Lc)$ with respect to the unknown true distribution $\PP$. To see this, assume that the type-1 Wasserstein distance between $\PP$ and its noisy estimate $\Pnom_N$ is known to be at most $\eps$. Thus, $\eps$ can be viewed as a measure of the estimation error. Assume further that a fixed loss function $\ell(\xi)$ is Lipschitz-continuous with Lipschitz constant $L$. The risk of $\ell$ then satisfies
\[
    \left| \risk(\Pnom_N, \ell)-\risk(\PP, \ell) \right| = L\cdot \left| \EE^{\Pnom_N}[\ell(\xi)/L] -  \EE^{\PP}[\ell(\xi)/L] \right| \le L\cdot W_1(\Pnom_N, \PP)\le L\cdot\eps,
\]
where the equality holds due to the definition of the risk, while the first inequality follows from the Kantorovich-Rubinstein theorem, which applies because $\Lip(\ell/L)\le 1$. Moreover, if all loss functions $\ell\in\Lc$ are Lipschitz continuous with the same Lipschitz constant $L$, then a similar reasoning implies that the optimal risk satisfies $| \risk(\Pnom_N, \Lc)-\risk(\PP, \Lc) | \le L\cdot \eps$. This analysis offers a rough understanding of how estimation errors in the input distribution are propagated to the (optimal) risk: they are amplified at most by the Lipschitz constant of the involved loss functions. Arguments of this type are central to the stability theory of stochastic programming. For example, it is known that under standard regularity conditions, the optimal values of two-stage stochastic programs with random right hand sides are Lipschitz continuous in the distribution of the uncertainty with respect to the Wasserstein distance~\cite{roemisch1991stability}. Classical stability results in stochastic programming are surveyed in~\cite{dupacova1990stability,roemisch2003}.

The above reasoning suggests that in order to approximate the (optimal) risk well, one should construct an estimator $\Pnom_N$ that has a small Wasserstein distance to the unknown true distribution $\PP$ with high confidence. Unfortunately, however, estimators are subject to fundamental performance limitations and cannot be improved beyond a certain level.

\begin{example}[Limitations of estimator performance]
    \label{ex:estimator-limitations}
    Depending on the available structural information on $\PP$, the nominal distributions portrayed in Example~\ref{ex:nominal-distribution}, which will be used throughout this tutorial, are essentially optimal within certain estimator families.
    \begin{enumerate}
        \item {\bf Discrete distributions:} Assume that $\PP$ is only known to be supported on a compact set $\Xi\subseteq \RR^m$, and let $\mathcal P_N$ be the family of all discrete distributions on $\Xi$ with $N$ atoms. The theory of optimal quantization shows that there exist $\underline N\in\mathbb N$ and $\underline c>0$ such that $\inf_{\QQ\in\mathcal P_N}W_1(\QQ,\PP)\ge \underline c N^{-1/m}$ for all $N\ge \underline N$ \cite[Theorem~3.3]{canas2012learning}. Thus, the type-1 Wasserstein distance between $\PP$ and its closest $N$-point distribution cannot decay faster than $N^{-1/m}$. Maybe surprisingly, the empirical distribution $\Pnom_N = \frac{1}{N} \sum_{i=1}^N \delta_{\widehat \xi_i}$ attains this optimal decay rate in a probabilistic sense even though it is constructed from $N$ random samples but without knowledge of $\PP$. Indeed, \cite[Theorem 2]{fournier2015rate} implies that for every $\eta\in(0,1)$  there exist $\overline N\in\mathbb N$ and $\overline c>0$ such that $W_1(\Pnom_N,\PP)\le \overline c N^{-1/m}$ with confidence $1-\eta$ for every $N\ge \overline N$. Thus, if we aim to approximate $\PP$ with a sequence of {\em discrete} distributions, the empirical distribution $\Pnom_N$ is essentially optimal in the sense that it attains the best possible convergence rate at any desired confidence~level.

        \item {\bf Elliptical distributions:} Assume that $\PP$ is known to be an elliptical distribution with a known density generator $g$ but unknown mean vector $\mu$ and covariance matrix~$\cov$. In this case, the problem of finding an estimator $\Pnom_N$ for the distribution $\PP$ reduces to finding an estimator $\widehat \theta_N$ for the vector $\theta=(\mu,\cov)$ of unknown distribution parameters. Under mild regularity conditions, the Cram\'er-Rao inequality guarantees that the covariance matrix of $\sqrt{N}\cdot \widehat\theta_N$ exceeds the inverse Fisher information matrix in a positive semidefinite sense for {\em any} unbiased estimator $\widehat \theta_N$. As the maximum likelihood estimator $\widehat \theta_N^{\, \rm ML}$ is asymptotically unbiased and efficient, {\em i.e.}, the mean of $\widehat\theta_N^{\, \rm ML}$ converges to~$\theta$ and the variance of $\sqrt{N}\cdot \widehat\theta_N^{\, \rm ML}$ converges to the inverse Fisher information matrix as $N$ grows, it is asymptotically optimal among all conceivable unbiased estimators. 
        \end{enumerate}
        We emphasize that, by mobilising more powerful results from statistics, the above optimality guarantees could be extended to even larger families of estimators. \hfill $\Box$
\end{example}

Example~\ref{ex:estimator-limitations} suggests that the accuracy of the nominal distribution cannot be increased beyond some fundamental limit by tuning the estimator. The only remaining option to reduce the estimation error is to increase the sample size~$N$, which may be expensive or impossible. Indeed, additional training samples may only become available in the future. Thus, the optimizer's curse illustrated in Example~\ref{ex:optimizers-curse} is fundamental and cannot be eliminated. However, once the potential to improve the estimator $\Pnom_N$ is exhausted, it may still be possible to mitigate the optimizer's curse by altering the risk evaluation and decision problems~\eqref{eq:risk} and~\eqref{eq:opt:risk} directly. Specifically, we propose here to robustify these problems against the uncertainty about the true distribution $\PP$. Distributional uncertainty is often referred to as {\em ambiguity} or {\em Knightian uncertainty} and is conveniently captured by an {\em ambiguity set}, that is, an uncertainty set in the space of probability distributions. To formalize this idea, we let $\Xi\subseteq\RR^m$ be a closed set that is known to contain the support of~$\PP$. In the absence of any structural information, we may simply set $\Xi=\RR^m$. Moreover, we denote by~$\mathcal P(\Xi)$ the family of all probability distributions supported on $\Xi$, and we define the ambiguity set
\begin{equation*}
\label{eq:ball}
\B_{\eps,p}(\Pnom_N) = \left\{ \QQ \in \mathcal P(\Xi): W_p(\QQ, \Pnom_N) \leq \eps \right\}
\end{equation*}
as the ball of radius $\eps\ge 0$ in $\mathcal P(\Xi)$ centered at the nominal distribution $\Pnom_N$ with respect to the type-$p$ Wasserstein distance. By construction, this ambiguity set contains all distributions supported on $\Xi$ that can be obtained by reshaping the nominal distribution~$\Pnom_N$ at a transportation cost of at most $\eps$. We can think of $\B_{\eps,p}(\Pnom_N)$ as the set of all distributions for which the estimation error---as measured by the type-$p$ Wasserstein distance---is at most $\eps$, and we can interpret $\eps$ as the maximum estimation error against which we seek protection.


Using the proposed ambiguity set, we define the {\em worst-case risk} as
\begin{equation}
\label{eq:worst:risk}
\risk_{\eps, p}(\Pnom_N, \ell) = \sup_{\QQ \in \B_{\eps,p}(\Pnom_N)} ~ \risk(\QQ, \ell)
\end{equation}
and the {\em worst-case optimal risk} as
\begin{equation}
\label{eq:dro}
\risk_{\eps, p}(\Pnom_N, \Lc) = \inf_{\ell \in \Lc} ~ \risk_{\eps, p}(\Pnom_N, \ell).
\end{equation}
Problem~\eqref{eq:dro} constitutes a distributionally robust optimization problem. It seeks decisions that have minimum risk under the most adverse distributions in the ambiguity set. Intuitively, problem~\eqref{eq:dro} can thus be viewed as a zero-sum game, where the decision-maker first selects an admissible loss function with the goal to minimize the risk, in response to which some fictitious adversary or `nature' selects a distribution from within the ambiguity set with the goal to maximize the risk. The hope is that by minimizing the worst-case risk, we actually push down the risk under {\em all} distributions in the ambiguity set---in particular under the unknown true distribution $\PP$, which is contained in the ambiguity set if $\eps$ is large enough. Thus, there is reason to hope that the solutions of distributionally robust optimization problems with carefully calibrated ambiguity sets display low out-of-sample~risk.


The distributionally robust risk evaluation and decision problems~\eqref{eq:worst:risk} and~\eqref{eq:dro} are attractive for a multitude of diverse reasons.

\begin{itemize}
\item {\bf Fidelity:} Distributionally robust models are more `honest' than their nominal counterparts as they acknowledge the presence of distributional uncertainty. They also benefit from information about the type and magnitude of the estimation errors, which is conveniently encoded in the geometry and size of the ambiguity set.
\item {\bf Managing expectations:} Due to the optimizer's curse, the solutions of nominal decision problems equipped with noisy estimators display an optimistic in-sample risk, which cannot be realized out of sample; see Example~\ref{ex:optimizers-curse}. In contrast, the solutions of distributionally robust decision problems are guaranteed to display an out-of-sample risk that falls below the worst-case optimal risk whenever the ambiguity set contains the unknown true distribution. Thus, nominal decision problems over-promise and under-deliver, while distributionally robust decision problems under-promise and over-deliver.
\item {\bf Computational tractability:}  The distributionally robust problems~\eqref{eq:worst:risk} and~\eqref{eq:dro} can often be reformulated as (or tightly approximated by) finite convex programs that are solvable in polynomial time. Section~\ref{sect:computation} will showcase some key tractability results.
\item {\bf Performance guarantees:} For judiciously calibrated ambiguity sets, one can prove that the worst-case optimal risk for any fixed sample size $N$ provides an upper confidence bound on the out-of-sample risk attained by the optimizers of~\eqref{eq:dro} (finite sample guarantee) and that the optimizers of~\eqref{eq:dro} converge almost surely to an optimizer of~\eqref{eq:opt:risk} as $N$ tends to infinity (asymptotic guarantee); see Section~\ref{sect:guarantees}.
\item {\bf Regularization by robustification:} The optimizer's curse is reminiscent of overfitting phenomena that plague most statistical learning models. One can show that distributionally robust learning models equipped with a Wasserstein ambiguity set are often equivalent to regularized learning models that minimize the sum of a nominal objective and a norm term that penalizes hypothesis complexity. Similarly, one can show that some distributionally robust maximum likelihood estimation models produce shrinkage estimators. Thus, Wasserstein distributional robustness offers new probabilistic interpretations for popular regularization techniques. The empirical success of regularization methods in statistics fuels hope that Wasserstein distributionally robust models can effectively combat the optimizer's curse across many application areas. Connections between robustification and regularization will be explored in Section~\ref{sect:app}.
\item {\bf Anticipating black swans:} If uncertainty is modeled by the empirical distribution, then the nominal decision problem evaluates the admissible loss functions only at the training samples. However, possible future uncertainty realizations that differ from all training samples but could have devastating consequences (`black swans') are ignored. If the empirical distribution may be perturbed within a Wasserstein ball with a positive radius, on the other hand, then (possibly small amounts of) probability mass can be moved anywhere in the support set~$\Xi$. Thus, the Wasserstein distributionally robust decision problem faithfully anticipates the possibility of black swans. We emphasize that all distributions in a Kullback-Leibler divergence ball must be absolutely continuous with respect to the nominal distribution, which implies that the corresponding distributionally robust decision problems ignore the possibility of black swans.
\item {\bf Axiomatic justification:} If the random vector $\xi$ may follow any distribution in some ambiguity set $\mathcal Q$ ({\em e.g.}, a Wasserstein ball), then the scalar random variable $\ell(\xi)$ corresponding to a fixed loss function $\ell\in\Lc$ may follow any distribution in the induced ambiguity set $\ell_*(\mathcal Q)=\{\ell_*(\QQ): \QQ\in\mathcal Q\}$, where $\ell_*(\QQ)$ is the pushforward measure of $\QQ$ under $\ell$. We call a loss function $\ell\in\Lc$ unambiguous if $\ell_*(\mathcal Q)$ is a singleton. Assume now that $\ell$ is preferred to $\ell'$ under any of the following natural conditions: (i)~$\ell$ and $\ell'$ are unambiguous, and $\risk(\QQ,\ell)\le\risk(\QQ,\ell')$ for some $\QQ\in\mathcal Q$; (ii)~$\ell_*(\mathcal Q)\subseteq \ell'_*(\mathcal Q)$; (iii)~$\risk(\QQ,\ell)\le \risk(\QQ,\ell')$ for every $\QQ\in\mathcal Q$. Under a mild technical condition, the loss functions must then be ranked by the worst-case risk $\sup_{\QQ\in\mathcal Q}\risk(\QQ,\ell)$ \cite[Theorem 12]{delage2019dicesion}. This result provides an axiomatic justification for adopting a distributionally robust approach.
\item {\bf Optimality principle:} Data-driven optimization aims to use the training data directly to construct an estimator for the objective of problem~\eqref{eq:opt:risk} (a predictor) and a decision that minimizes this predictor (a prescriptor) without the detour of constructing an estimator for $\PP$. It has been shown that optimal predictors and the corresponding prescriptors can be constructed by solving a meta-optimization model that minimizes the in-sample risk of the predictor-prescriptor pairs subject to constraints guaranteeing that the in-sample risk is actually attainable out of sample. It has been shown that this meta-optimization problem admits a unique solution: the best predictor-prescriptor pair is obtained by solving a distributionally robust optimization problem over all distributions in some neighborhood of the empirical distribution \cite[Theorem~7]{vanparys2019optimal}. Thus, if one aims to transform training data to decisions, it is in some precise sense optimal to do this by solving a data-driven distributionally robust optimization problem. 
\end{itemize}

Distributionally robust optimization models with Wasserstein ambiguity sets were introduced in~\cite{pflug2007ambiguity}. Reformulations of these models as nonconvex optimization problems as well as initial attempts to solve these problems via algorithms from global optimization are reported in \cite{wozabal2012framework} and \cite[\S~7.1]{pflug2014multistage}. In the next section we will review convex reformulations and approximations that were discovered in~\cite{esfahani2018data, zhao2018data} and significantly generalized in~\cite{blanchet2016quantifying, gao2016distributionally}.

\vspace{1mm}

\paragraph{Notation.}
The conjugate of a function $\ell(\xi)$ on $\RR^m$ is defined as $\ell^*(z) = \sup_{\xi} z^\top \xi - \ell(\xi)$. The indicator function of a set $\Xi \subseteq \RR^m$ is defined as $\delta_\Xi(\xi) = 0$ if $\xi \in \Xi$ and $\delta_\Xi(\xi)= \infty$ if $\xi\notin\Xi$. The conjugate $\sigma_\Xi(z) = \sup_{\xi \in \Xi}  z^\top \xi$ of the indicator function is termed the support function. If $ \| \xi \| $ represents the norm of~$\xi\in\RR^m$, then $ \| z \|_* =\sup_{\| \xi \| \leq 1} z^\top \xi$ denotes the corresponding dual norm. The set of all symmetric (positive semidefinite) matrices $A \in \RR^{m \times m}$ is denoted by~$\mathbb{S}^m$ ($\PSD^m$). For $A,B\in\mathbb S^m$, the relation $A\succeq B$ ($A\succ B$) means that $A-B$ is positive semidefinite (positive definite). The trace of $A\in \RR^{m \times m}$ is denoted by $\Tr{A}$, the smallest and largest eigenvalues of $A \in \mathbb{S}^m$ are denoted by $\eig_{\min}(A)$ and $\eig_{\max}(A)$, respectively, and the Moore-Penrose pseudoinverse of $A \in \PSD^m$ is denoted by $A^\dagger$. For $N \in \mathbb{N}$ we set $[N] = \{1, \ldots, N\}$.

\section{Computation}
\label{sect:computation}
The aim of this section is to show that the worst-case risk evaluation problem~\eqref{eq:worst:risk} and the distributionally robust decision problem~\eqref{eq:dro} are computationally tractable in many situations of practical interest. Note first that checking whether a fixed distribution $\QQ$ is feasible in~\eqref{eq:worst:risk} requires computing the Wasserstein distance $W_p(\QQ,\Pnom_N)$. It is therefore instructive to study the complexity of evaluating Wasserstein distances between arbitrary distributions.

Computing the Wasserstein distance between two discrete distributions amounts to solving a tractable linear program that is susceptible to the network simplex algorithm~\cite{bertsekas1998network} as well as dual ascent methods~\cite{bertsimas1997introduction} or specialized auction algorithms~\cite{bertsekas1981new, bertsekas1992auction}, etc. The set of feasible transportation plans is termed the transportation polytope and displays many useful theoretical properties, which are surveyed in~\cite{brualdi2006combinatorial}. The need to evaluate Wasserstein distances between increasingly fine-grained histograms has recently motivated efficient approximation schemes. When augmented with an entropic regularization term, for instance, the finite-dimensional transportation problem can be solved quickly by using Sinkhorn's algorithm~\cite{chizat2018scaling, cuturi2013sinkhorn, karlsson2017generalized, peyre2015entropic, peyre2017quantum, schmitzer2016sparse, solomon2015convolutional}. Variants of this approach use Tikhonov regularizers~\cite{essid2018quadratically}, Bregman divergences~\cite{benamou2015iterative} or Tsallis entropies~\cite{muzellec2017tsallis} instead of the entropic regularization term. A survey of algorithms for the finite-dimensional transportation problem is provided in~\cite{peyre2018computational}.

As soon as at least one of the two involved distributions ceases to be discrete, the Wasserstein distance can no longer be evaluated in polynomial time. Even in the simplest imaginable scenario where one distribution is uniform on a hypercube and the other distribution is discrete with two atoms, computing the Wasserstein distance becomes intractable \cite{taskesen2019complexity}.

\begin{theorem}[Hardness of computing Wasserstein distances]
    \label{theorem:hard}
    Computing the type-$p$ Wasserstein distance between two distributions $\QQ$ and $\QQ'$ is \#P-hard even if $\|\cdot\|$ is the Euclidean norm, $\QQ$ is the uniform distribution on the standard hypercube $[0,1]^m$, and $\QQ'$ is a discrete distribution supported on only two points. 
\end{theorem}

If $p=2$ and $\|\cdot\|$ is the Euclidean norm, then the Wasserstein distance admits an analytical lower bound that depends only on the distributions' first- and second-order moments. This bound is available for {\em any} pair of distributions even if their exact Wasserstein distance cannot be computed efficiently. Moreover, the bound is exact for elliptical distributions.

\begin{theorem}[Gelbrich bound]
    \label{theorem:gelbrich}
    If $\|\cdot\|$ is the Euclidean norm, and the distributions $\QQ$ and $\QQ'$ have mean vectors $\mu, \mu' \in \RR^m$ and covariance matrices $\Sigma, \Sigma' \in \PSD^m$, respectively, then
    \begin{equation}
    \label{eq:gelbrich-bound}
    W_2(\QQ, \QQ') \geq \sqrt{\| \mu - \mu' \|_2^2 + \Tr{\Sigma + \Sigma' - 2 \left( \Sigma^{\frac{1}{2}} \Sigma' \Sigma^{\frac{1}{2}} \right)^{\frac{1}{2}} } }.
    \end{equation}
    The bound is exact if $\QQ$ and $\QQ'$ are elliptical distributions with the same density generator.
\end{theorem}

The inequality~\eqref{eq:gelbrich-bound} may be loose if $\QQ$ and $\QQ'$ are elliptical distributions with different density generators. Maybe unexpectedly, however, the Wasserstein distance between two elliptical distributions with the same density generator $g$ is actually independent of $g$. In its general form, Theorem~\ref{theorem:gelbrich} is due to Gelbrich~\cite{gelbrich1990formula}. The exact formula for the type-2 Wasserstein distance between normal distributions has been discovered earlier in \cite{dowson1982frechet, givens1984class, olkin1982distance}.

As any Wasserstein ball with a strictly positive radius contains non-discrete distributions (the nominal distribution can be smeared out even if the transportation budget is small), it is perhaps surprising that the worst-case risk evaluation problem~\eqref{eq:worst:risk} may be tractable at all. Indeed, Theorem~\ref{theorem:hard} indicates that checking feasibility is already hard in general. We will see below, however, that the extremal distributions determining the worst-case risk are often structurally equivalent to the nominal distribution. Thus, there is hope that problems~\eqref{eq:worst:risk} and~\eqref{eq:dro} become tractable if we choose a nominal distribution with a particularly simple structure ({\em e.g.}, a discrete or an elliptical distribution).

In order to ensure that any admissible loss function $\ell\in\Lc$ has a finite expected value under the nominal distribution, we impose the following technical regularity condition borrowed from~\cite{blanchet2016quantifying}, which will tacitly be assumed to hold throughout the rest of the paper.

\begin{assumption}[Regularity]
    \label{assumption:loss}
    Any loss function $\ell \in \Lc$ is upper semicontinuous and integrable with respect to the nominal distribution $\Pnom_N$, that is, $ \int_{\RR^m} |\ell(\xi)| \, \Pnom_N(\diff \xi) < \infty. $
\end{assumption}

In the remainder of this section, we will first review tractable bounds on the worst-case risk and present a strong duality result that paves the way towards exact tractable reformulations~(Section~\ref{sec:wc-risk-any-p}). Next, we will delineate efficient methods to compute the worst-case risk as well as the underlying worst-case distributions in situations when the nominal distribution is discrete (Section~\ref{sec:wc-risk-empirical}) or elliptical (Section~\ref{sec:wc-risk-elliptical}).

\subsection{General Analysis of the Worst-Case Risk}
\label{sec:wc-risk-any-p}
Before attempting to derive exact tractable reformulations for the worst-case risk~\eqref{eq:worst:risk}, we focus on the simpler task of establishing efficiently computable upper and lower bounds. To derive a pessimistic upper bound, we note that the transportation cost $\|\xi-\xi'\|^p$ is a convex function of the random variable $\|\xi-\xi'\|$ for any $p\ge 1$. Jensen's inequality thus implies
\[
    W_p(\QQ, \Pnom_N) \geq W_1(\QQ, \Pnom_N) \quad \forall \QQ\in\mathcal P(\Xi)\quad \implies \quad \B_{\eps,p}(\Pnom_N) \subseteq  \B_{\eps,1}(\Pnom_N).
\]
Hence, the worst-case risk of a loss function $\ell\in\Lc$ over the type-$p$ Wasserstein ball satisfies
\begin{align*}
    \risk_{\eps, p}(\Pnom_N, \ell) \leq \risk_{\eps, 1}(\Pnom_N, \ell) &= \EE^{\Pnom_N}[\ell(\xi)] + \sup_{\QQ\in \B_{\eps,1}(\Pnom_N) } \EE^{\QQ}[\ell(\xi)] - \EE^{\Pnom_N}[\ell(\xi)] \\
    & \le \risk(\Pnom_N, \ell) + \eps\cdot \Lip(\ell),
\end{align*}
where the equality follows from the definition of the worst-case risk, while the second inequality is a direct consequence of the Kantorovich-Rubinstein theorem (see Theorem~\ref{thm:kantorovich-rubinstein}). We summarize the above reasoning in the following theorem.

\begin{theorem}[Lipschitz regularization]
    \label{theorem:lipschitz}
    The worst-case risk~\eqref{eq:worst:risk} of any fixed loss function $\ell\in\Lc$ is bounded above by the Lipschitz-regularized nominal risk, that is,
    \begin{equation*}
        \risk_{\eps, p}(\Pnom_N, \ell) \leq \risk(\Pnom_N, \ell) + \eps\cdot \Lip(\ell).
    \end{equation*}
\end{theorem}

If the loss function $\ell$ fails to be Lipschitz continuous ({\em i.e.}, $\Lip(\ell) = \infty$), then Theorem~\ref{theorem:lipschitz} is trivially satisfied. Note that $\risk(\Pnom_N, \ell)$ is linear in $\ell$ for any choice of the nominal distribution and that $\Lip(\ell)$ is a convex function of $\ell$. Thus, minimizing the upper bound of Theorem~\ref{theorem:lipschitz} amounts to solving a convex optimization problem whenever $\Lc$ is a convex set.



An optimistic lower bound on the worst-case risk can be obtained by replacing the Wasserstein ball in~\eqref{eq:worst:risk} with a smaller ambiguity set. If the distributions in the restricted Wasserstein ball admit a finite parameterization, then the lower bounding problem coincides with a finite optimization problem. Depending on the parameterization, this problem may even be convex. If $\Pnom_N$ is the empirical distribution, for example, one may restrict the original Wasserstein ball to a subset that contains only {\em perturbed} empirical distributions of the form
\[
    \QQ(\Theta) = \frac{1}{N} \sum_{i=1}^N \delta_{\widehat \xi_i+\theta_i},
\]
where $\theta_i\in\RR^m$ is the displacement of the $i^{\rm th}$ training sample. Thus, all distributions in the restricted Wasserstein ball are encoded by a perturbation matrix $\Theta=(\theta_1,\ldots, \theta_N)\in\RR^{m\times N}$. The requirement that $\QQ(\Theta)\in\mathcal P(\Xi)$ translates to $\widehat \xi_i + \theta_i \in \Xi$ for all $i\in[N]$, while the Wasserstein constraint $W_p(\QQ(\Theta),\Pnom_N)\le \eps$ is equivalent to the inequality $\frac{1}{N} \sum_{i=1}^N \| \theta_i \|^p \leq \eps^p$.

\begin{theorem}[Robust lower bound]
\label{thm:robust-lb}
    If $\Pnom_N$ is the empirical distribution, then the worst-case risk~\eqref{eq:worst:risk} of any fixed loss function $\ell\in\Lc$ is bounded below by the worst-case empirical loss, where the worst case is taken over all perturbation matrices $\Theta=(\theta_1,\ldots, \theta_N)\in\RR^{m\times N}$ of the training samples in an $L_{p,1}$-norm uncertainty set, that is, we have
    \begin{equation}
    \label{eq:robust}    
        \optimize{
            \risk_{\eps, p}(\Pnom_N, \ell) ~\geq ~\sup & \DS \frac{1}{N} \sum_{i=1}^N \ell(\widehat \xi_i + \theta_i) \\ [3ex]
            \hspace{6.8em}\st & \theta_i \in \RR^m & \forall i \in [N] \\ [0.5em]
            & \widehat \xi_i + \theta_i \in \Xi & \forall i \in [N] \\ 
            & \DS \frac{1}{N} \sum_{i=1}^N \| \theta_i \|^p \leq \eps^p.
        }
    \end{equation}
\end{theorem}

Note that if the loss function $\ell$ is concave, then the robust lower bounding problem of Theorem~\ref{thm:robust-lb} constitutes a finite convex optimization problem. In the remainder we will argue that both the upper bound of Theorem~\ref{theorem:lipschitz} as well as the lower bound of Theorem~\ref{thm:robust-lb} can become exact in situations of practical interest. To see this, we first derive the Lagrangian dual of the worst-case risk evaluation problem~\eqref{eq:worst:risk}.

\begin{theorem}[Strong duality]
    \label{theorem:robust}
    The worst-case risk~\eqref{eq:worst:risk} of any fixed $\ell\in\Lc$ satisfies
    \begin{equation}
        \label{eq:strong-duality}
        \risk_{\eps, p}(\Pnom_N, \ell) = \inf_{\gamma \ge 0} ~ \EE^{\Pnom_N} \left[ \ell_\gamma(\xi) \right] + \gamma \eps^p,
    \end{equation}
    where $ \ell_\gamma(\xi) = \sup_{z \in\Xi} ~ \ell(z) - \gamma \| z - \xi \|^p$ is a Moreau-Yosida regularization~\cite{parikh2014proximal} of $\ell(\xi)$.
\end{theorem}

The minimization problem on the right hand side of~\eqref{eq:strong-duality} can indeed be identified with the strong Lagrangian dual of problem~\eqref{eq:worst:risk}, where $\gamma\ge 0$ is the Lagrange multiplier of the Wasserstein constraint $W_p(\QQ,\Pnom_N)\le \eps$. We emphasize that the Moreau-Yosida regularization $\ell_\gamma(\xi)$ is jointly convex in $\gamma$ and $\ell$ for every fixed uncertainty realization $\xi$. Thus the dual problem~\eqref{eq:strong-duality} represents a convex minimization problem whose optimal value is convex in~$\ell$. One can further show that~\eqref{eq:strong-duality} is  solvable for any $\eps>0$ under the mild assumption that there exists $C>0$ such that $|\ell(\xi)|\le C(1+\|\xi\|^p)$ for all $\xi\in\Xi$. For type-1 Wasserstein balls centered at the empirical distribution, Theorem~\ref{theorem:robust} is a corollary of \cite[Theorem~4.2]{esfahani2018data} and \cite[Proposition~2]{zhao2018data}. An extension of Theorem~\ref{theorem:robust} to situations where~$\xi$ ranges over a Polish space is discussed in \cite{blanchet2016quantifying, gao2016distributionally}. It has been shown that Theorem~\ref{theorem:robust} remains even valid if the transportation cost $\|\xi-\xi'\|^p$ in the definition of the Wasserstein distance is replaced with a general nonnegative and lower semicontinuous function $c(\xi, \xi')$ that vanishes if and only if~$\xi = \xi'$ \cite{blanchet2016quantifying}. Note that the Wasserstein distance may cease to be a metric in this case.


In the next sections we will describe specific settings in which~\eqref{eq:worst:risk} and~\eqref{eq:strong-duality} are tractable.


\subsection{Tractability Results for Empirical Nominal Distributions}
\label{sec:wc-risk-empirical}
Assume now that the Wasserstein ambiguity set is centered at the empirical distribution defined in~\eqref{eq:empirical}. In this case, under a mild convexity assumption, the worst-case risk~\eqref{eq:worst:risk} can be exactly expressed as the optimal value of a finite convex optimization problem.

\begin{theorem}[Piecewise concave loss I]
    \label{thm:type1}
    Assume that $\Xi$ is convex and closed and that $\ell(\xi) = \max_{j \in [J]} \ell_j(\xi)$, where $-\ell_j$ is proper, convex and lower semicontinuous for all $j\in[J]$. If $\Pnom_N$ is the empirical distribution and $p,q\ge 1$ with $\frac{1}{p}+\frac{1}{q}=1$, then the worst-case risk~\eqref{eq:worst:risk} coincides with the optimal value of a finite convex minimization problem, that is, 
    \begin{align}
        \label{eq:peyman}
        &\risk_{\eps, p}(\Pnom_N, \ell)~ = \nonumber \\  
        &\begin{array}{c@{~~~}l@{~~}ll}
            \inf  & \DS \gamma \eps^p + \frac{1}{N} \sum_{i=1}^N s_i \\ [1em]
            \st \!\!\! & \DS \gamma\in\RR_+,~ s_i\in\RR,~ u_{ij} \in\RR^m,~ v_{ij}\in\RR^m & \forall i\in[N],\; j \in [J] \\
            & \DS [-\ell_j]^* (u_{ij} - v_{ij}) + \sigma_\Xi(v_{ij}) - u_{ij}^\top \widehat \xi_i + \varphi(q) \,\gamma \left\| \frac{u_{ij}}{\gamma} \right\|_*^q \leq s_i &\forall i\in[N],\; j \in [J],
        \end{array}
    \end{align}
    where $[-\ell_j]^*(z)$ is the conjugate of $-\ell_j(\xi)$, $\sigma_\Xi(z)$ is the support function of $\Xi$, and $\| \cdot \|_*$ is the dual of the norm $\|\cdot\|$ on $\RR^m$, while $\varphi(q)=(q-1)^{q-1}/q^q$ for $q>1$ and $\varphi(1)=1$. For $\gamma=0$, the expression $0 \|u_{ij}/0\|_*^q$ is interpreted as $\lim_{\gamma\downarrow 0} \gamma \|u_{ij}/\gamma\|^q_*$.
\end{theorem}

The assumptions of Theorem~\ref{thm:type1} are unrestrictive because any continuous function $\ell(\xi)$ on a compact set $\Xi$ can be uniformly approximated as closely as desired by a pointwise maximum of finitely many concave functions. Note that the loss function~$\ell(\xi)$ and the support set~$\Xi$ enter problem~\eqref{eq:peyman} through the conjugates of the negative constituent functions $-\ell_j(\xi)$, $j\in[J]$, and the support function~$\sigma_\Xi(z)$, all of which are convex. Moreover, the norm that determines the transportation cost in the definition of the Wasserstein distance enters~\eqref{eq:peyman} via the dual norm~$\|\cdot\|_*$. The term $\gamma \| u_{ij}/\gamma \|_*^q$ can be identified with the perspective function of $\|u_{ij}\|_*^q$ and is thus jointly convex in $\gamma$ and $u_{ij}$ \cite[\S~3.2.6]{boyd2004convex}. Therefore, problem~\eqref{eq:peyman} is manifestly convex. Tables~\ref{table:conjugate}--\ref{table:norm} in Appendix~\ref{sect:functions} list common conjugates, support functions and dual norms. By substituting~\eqref{eq:peyman} into~\eqref{eq:dro}, one can reformulate the distributionally robust decision problem~\eqref{eq:dro} as a single explicit minimization problem, which is convex whenever $\Lc$ is a convex set. To prove Theorem~\ref{thm:type1}, one re-expresses the empirical expectation in the objective function of the dual problem~\eqref{theorem:robust} as a finite sum and dualizes the maximization problems in the Moreau-Yosida regularization terms. For further details see~\cite[Theorem~4.2]{esfahani2018data} and \cite{zhen2019distributionally}.

\begin{remark}[Limiting cases I]
\label{rmk:lim}
In the limit when $p$ tends to 1 and $q$ to $\infty$, the function $\varphi(q)$ decays as $1/q$, while $\|u_{ij}/\gamma\|_*^q$ grows exponentially whenever $\|u_{ij}\|_*>\gamma$. Thus, we have 
\[
    \lim_{q\uparrow \infty}~ \varphi(q) \,\gamma \left\| \frac{u_{ij}}{\gamma} \right\|_*^q = \left\{ \begin{array}{c@{\quad}l}
    0 & \text{if }\|u_{ij}\|_*\le \gamma, \\
    \infty & \text{if }\|u_{ij}\|_*>\gamma.
    \end{array}\right.
\]
For $p=1$, the constraints of the finite convex program~\eqref{eq:peyman} are thus equivalent to
\[
    \DS [-\ell_j]^* (u_{ij} - v_{ij}) + \sigma_\Xi(v_{ij}) - u_{ij}^\top \widehat \xi_i  \leq s_i,\quad \|u_{ij}\|_*\le \gamma \quad \forall i\in[N],\; j \in [J].
\]
In the opposite limit when $p$ tends to $\infty$ and $q$ to 1, the function $\varphi(q)$ converges to 1. One can then show that $\gamma=0$ at optimality and that the constraints of problem~\eqref{eq:peyman} simplify~to\footnote{Updated.}
\[
    \DS [-\ell_j]^* (u_{ij} - v_{ij}) + \sigma_\Xi(v_{ij}) - u_{ij}^\top \widehat \xi_i + \eps \left\| u_{ij} \right\|_* \leq s_i \quad \forall i\in[N],\; j \in [J].
\] 
We refer to \cite[Appendix~A.6]{wang2022mean} for a formal proof.
\hfill $\Box$
\end{remark}

As it expresses the worst-case risk $\risk_{\eps, p}(\Pnom_N, \ell)$ as the optimal value of a {\em minimization} problem, Theorem~\ref{thm:type1} primarily serves as a vehicle to solve the distributionally robust decision problem~\eqref{eq:dro}. In order to construct an extremal distribution that solves problem~\eqref{eq:worst:risk}, one may dualize the finite convex program~\eqref{eq:peyman} to convert it back to a {\em maximization} problem.

\begin{theorem}[Piecewise concave loss II]
    \label{thm:type1:extrem}
    Under the conditions of Theorem~\ref{thm:type1} we have
    \begin{equation}
        \label{dual-dual}
        \begin{array}{r@{\quad}l@{\quad~}l}
            \risk_{\eps, p}(\Pnom_N, \ell) ~=~ \max & \DS \frac{1}{N} \sum_{i=1}^N \sum_{j=1}^J \alpha_{ij} \ell_j \Big( \widehat \xi_i + \frac{\theta_{ij}}{\alpha_{ij}} \Big) \\ [4ex]
            \st & \alpha_{ij} \in \RR_+,\; \theta_{ij} \in \RR^m & \forall i \in [N],~ \forall j \in [J] \\ [1ex]
            & \DS \widehat \xi_i + \frac{\theta_{ij}}{\alpha_{ij}} \in \Xi & \forall i \in [N],~ \forall j \in [J] \\        
            & \DS \sum_{j=1}^J \alpha_{ij} = 1 & \forall i \in [N] \\
            & \DS \frac{1}{N} \sum_{i=1}^N \sum_{j=1}^J \alpha_{ij} \left\| \frac{\theta_{ij}}{\alpha_{ij}} \right\|^p \leq \eps^p,
        \end{array}
    \end{equation}
    where $0\, \ell_j( \widehat \xi_i + \theta_{ij}/0)$ is defined as the value that makes the function $\alpha_{ij} \ell_j( \widehat \xi_i + \theta_{ij}/\alpha_{ij})$ upper semicontinuous at $(\theta_{ij},\alpha_{ij})= (\theta_{ij}, 0)$. Similarly, the constraint $\widehat \xi_i + \theta_{ij}/0 \in \Xi$ means that $\theta_{ij}$ belongs to the recession cone of $\Xi$, and $0 \|\theta_{ij}/0\|^p$ is interpreted as $\lim_{\alpha_{ij}\downarrow 0} \alpha_{ij} \| \theta_{ij}/\alpha_{ij}\|^p$.
\end{theorem}


Problem~\eqref{dual-dual} is the Lagrangian dual of~\eqref{eq:peyman} and thus convex by construction. Convexity can also be verified directly. As the constituent functions $\ell_j(\xi)$, $j\in[J]$, are concave by assumption, the objective of~\eqref{dual-dual} represents a sum of concave perspective functions and is thus concave \cite[\S~3.2.6]{boyd2004convex}. Also, the support constraints $\widehat \xi_i +\theta_{ij}/\alpha_{ij}\in \Xi$ require that~$(\theta_{ij}, \alpha_{ij})$ belongs to the preimage of the convex set $\{\widehat \xi_i +\xi:\xi\in \Xi\}$ under the perspective transformation, which is known to be convex \cite[\S~2.3.3]{boyd2004convex}. The term $\alpha_{ij} \| \theta_{ij}/\alpha_{ij} \|^p$ can be identified with the perspective function of $\|\theta_{ij}\|^p$ and is thus jointly convex in $\alpha_{ij}$ and $\theta_{ij}$~\cite[\S~3.2.6]{boyd2004convex}.


For a proof of Theorem~\ref{thm:type1:extrem} we refer to \cite[Theorem~4.4]{esfahani2018data} and \cite{zhen2019distributionally}. We emphasize that problem~\eqref{dual-dual} is always solvable because it has a compact feasible set and an upper semicontinuous objective function, and thus the use of the maximization operator is justified.

Note that if $J=1$, then the loss function $\ell(\xi)=\ell_1(\xi)$ is globally concave, and the penultimate constraint group in~\eqref{dual-dual} simplifies to the requirement that $\alpha_{i1}=1$ for every $i\in[N]$. In this case, the convex program~\eqref{dual-dual}, which is equivalent to the worst-case risk evaluation problem~\eqref{eq:worst:risk}, reduces to the robust optimization problem~\eqref{eq:robust}, which maximizes only over perturbed empirical distributions in the Wasserstein ball. Thus, the robust lower bound portrayed in Theorem~\ref{thm:robust-lb} is exact if the loss function $\ell(\xi)$ is concave.

\begin{remark}[Limiting cases II]
For $p=1$, the last constraint of~\eqref{dual-dual} simplifies to
\[
    \DS \frac{1}{N} \sum_{i=1}^N \sum_{j=1}^J \left\| \theta_{ij}\right\| \leq \eps.
\]
To analyze the limit when $p$ tends to $\infty$, we divide the last constraint of~\eqref{dual-dual} by $\eps^p$ and observe that $\| \theta_{ij}/(\eps \, \alpha_{ij}) \|^p$ grows exponentially with $p$ if $\|\theta_{ij}/\alpha_{ij}\|> \eps$. Otherwise, $\| \theta_{ij}/(\eps \, \alpha_{ij}) \|^p$ remains bounded by 1 for all $p$. For $p=\infty$, the  last constraint of~\eqref{dual-dual} is therefore equivalent to the requirement that $\|\theta_{ij}\|\le \eps \,\alpha_{ij}$ for all $i\in [N]$ and $j\in[J]$. 

An intimate connection between distributionally robust optimization with type-$\infty$ Wasserstein balls and classical robust optimization has first been discovered in~\cite{sturt2018data-driven}. \hfill $\Box$
\end{remark}

Even though problem~\eqref{dual-dual} is guaranteed to have an optimal solution, the worst-case risk~\eqref{eq:worst:risk} may not be attained by any distribution if $p=1$. An instance of problem~\eqref{eq:worst:risk} that fails to be solvable is constructed in Example~\ref{eq:non-existence} below, which replicates~\cite[Example~2]{esfahani2018data}.

\begin{example}[Non-existence of extremal distributions]
\label{eq:non-existence}
Assume that $p=1$, $\Xi = \RR$, $N = 1$ and $\widehat \xi_1 = 0$ implying that the nominal distribution $\Pnom_1$ reduces to the Dirac distribution at~$0$. Set the norm on $\RR$ to the absolute value $|\cdot|$, and set $\ell(\xi ) = \max \{ 0, \xi-1 \}$. As $\Lip(\ell)=1$, Theorem~\ref{theorem:lipschitz} implies that $\risk_{\eps, 1}(\Pnom_1, \ell) \le \eps$. Next, define $\QQ_n = ( 1 - 1 /n ) \, \delta_{0} + (1 / n) \, \delta_{\eps n} $ for $n \in \mathbb N$, and note that the type-1 Wasserstein distance between $\QQ_n$ and $\Pnom_1$ amounts to~$\eps$, which is the cost of moving mass $1/n$ from $\eps n$ to $0$. Thus, $\QQ_n\in\B_{\eps,1}(\Pnom_1)$. Moreover, we have $\EE^{\QQ_n}[\ell(\xi)]=\max\{0,\eps-1/n\}$, which implies that $\QQ_n$ attains the upper bound $\eps$ on the worst-case risk asymptotically as $n$ tends to infinity. Thus, the sequence $\QQ_n$, $n\in\mathbb N$, is asymptotically optimal in~\eqref{eq:worst:risk}. Next, we argue that the worst-case risk $\eps$ is not attained. Suppose to the contrary that there exists $\QQ\opt \in \B_{\eps, 1}(\Pnom_1)$ with $\EE^{\QQ\opt}[\ell(\xi)] = \eps$. Thus, $\eps = \EE^{\QQ\opt}[\ell(\xi)] < \EE^{\QQ\opt}[|\xi|] \leq \eps$ where the strict inequality follows from the observation that $\ell(\xi) < |\xi|$ for any $\xi \neq 0$ and that $\QQ\opt \neq \delta_{0}$, and the second inequality follows from Theorem~\ref{thm:kantorovich-rubinstein} and the assumption that $\QQ\opt \in \B_{\eps, 1}(\Pnom_1)$. The contradiction implies that $\QQ\opt$ cannot exist, and thus~\eqref{eq:worst:risk} is not solvable. \hfill $\Box$
\end{example}

Fix now any maximizer $\{\alpha_{ij}\opt,\theta_{ij}\opt\}_{i,j}$ of problem~\eqref{dual-dual}. This maximizer can be used to construct an extremal distribution $\QQ\opt$ that solves problem~\eqref{eq:worst:risk} (if such a $\QQ\opt$ exists) or a sequence of asymptotically optimal distributions $\{\QQ_n\}_{n\in\mathbb N}$ (if such a $\QQ\opt$ does not exist). Before describing this construction, we remark that $\theta_{ij}\opt$ is a recession direction of the support set~$\Xi$ whenever $\alpha_{ij}\opt=0$ ({\em i.e.}, $\widehat\xi_i+t \, \theta_{ij}\opt\in\Xi$ for every $t\ge 0$). If $\Xi$ is bounded, this implies that $\theta_{ij}\opt=0$ whenever $\alpha_{ij}\opt=0$. Next, define $\nu_+$ as the set of pairs $(i,j)$ with $\alpha_{ij}\opt>0$, $\nu_0$ as the set of pairs $(i,j)$ with $\alpha_{ij}\opt=0$ and $\theta_{ij}\opt=0$, and $\nu_\infty$ as the set of pairs $(i,j)$ with $\alpha_{ij}\opt=0$ and $\theta_{ij}\opt\neq0$. By construction, $\nu_+$, $\nu_0$ and $\nu_\infty$ form a partition of $[N]\times [J]$.

If $\nu_\infty=\emptyset$, one can show that 
    \begin{equation*}
        \QQ\opt = \DS \sum_{(i,j)\in\nu_+} \!\!\! \frac{ \alpha_{ij}\opt}{N} \; \delta_{\widehat \xi_i + \theta\opt_{ij}/\alpha_{ij}\opt} 
    \end{equation*}        
is an extremal distribution that solves~\eqref{eq:worst:risk}. For $p>1$, the last constraint in~\eqref{dual-dual} ensures that~$\theta_{ij}=0$ whenever $\alpha_{ij}=0$ because otherwise $\alpha_{ij}\| \theta_{ij}/\alpha_{ij}\|^p$ evaluates to $\infty$. This implies that the set $\nu_\infty$ is empty. Thus, for $p>1$, the worst-case risk~\eqref{eq:worst:risk} of a piecewise concave loss function is always attained by the discrete distribution $\QQ\opt$ constructed above.

If $\nu_\infty\neq\emptyset$, which is only possible in the special case $p=1$, the distributions
    \begin{equation*}
    \QQ_n = \DS \sum_{(i,j)\in\nu_+\cup \nu_\infty} \!\!\!  \frac{\alpha_{ij}(n)}{N}  \; \delta_{\widehat \xi_i + \theta_{ij}\opt/\alpha_{ij}(n)}
    \quad \text{with} \quad
        \alpha_{ij}(n)=\left\{ \begin{array}{c@{~~}l}
    \alpha_{ij}\opt \left(1-\frac{|\nu_\infty|}{n}\right)& \text{if } (i,j)\in\nu_+,\\
    \frac{1}{n}  & \text{if } (i,j)\in\nu_\infty,
    \end{array}\right.
    \end{equation*}        
are feasible and asymptotically optimal in~\eqref{eq:worst:risk} as $n\ge |\nu_\infty|$ tends to infinity. Intuitively, these distributions send some atoms with decaying probabilities to infinity along specific recession directions $\theta_{ij}\opt$, $(i,j)\in\nu_\infty$, of the support set. Note that moving an atom to infinity is possible even when only a finite (type-1) transportation budget is available provided that the probability mass transported is inversely proportional to the transportation distance. 

For $p>1$, atoms can also migrate to infinity at a finite transportation cost provided that their probabilities are inversely proportional to the $p^{\rm th}$ power of the transportation distance. As piecewise concave loss functions grow at most linearly, however, the decay in probability always outweighs the increase in loss. This reasoning provides an intuitive explanation for our insight that $\nu_\infty=\emptyset$ and that the supremum in~\eqref{eq:worst:risk} is always attained for~$p>1$.


Given the promising results for piecewise concave loss functions, it is natural to ask whether the convex reformulations of Theorems~\ref{thm:type1} and~\ref{thm:type1:extrem} can be generalized. Indeed, it has been discovered that similar results are available for convex (but not piecewise convex) loss functions under the additional condition that there are no support constraints ($\Xi=\RR^m$).

\begin{theorem}[Convex loss and $p=1$]
    \label{thm:type1:convex}
    Assume that $ \Xi = \RR^m$ and that the loss function $\ell(\xi)$ is convex. If $p=1$ and $\Pnom_N$ is the empirical distribution, then the worst-case risk~\eqref{eq:worst:risk} coincides with the Lipschitz-regularized empirical loss, that is,
    \begin{equation*}
        \risk_{\eps, 1}(\Pnom_N, \ell) = \risk(\Pnom_N , \ell) + \eps \Lip(\ell).
    \end{equation*}
\end{theorem}

For a proof of this result we refer to \cite[Theorem~6.3]{esfahani2018data}. Theorem~\ref{thm:type1:convex} shows that the simple upper bound of Theorem~\ref{theorem:lipschitz} is exact if $p=1$, $\Xi=\RR^m$ and the loss function $\ell$ is convex.

\begin{remark}[Computing the Lipschitz modulus]
\label{rem:lipschitz}
By Theorem~\ref{thm:type1:convex}, computing the worst-case risk of a convex loss function $\ell(\xi)$ requires computing the Lipschitz modulus of~$\ell(\xi)$ with respect to the prescribed norm $\|\cdot\|$ on $\RR^m$. One can show that
\begin{equation}
    \label{eq:lipschitz-computation}
    \Lip(\ell) = \sup \left\{ \|z\|_* : \ell^*(z)<\infty \right\},
\end{equation}
that is, the Lipschitz modulus of $\ell(\xi)$ coincides with the radius of the smallest dual norm ball around 0 that encloses the domain of the conjugate loss function $\ell^*(z)$ \cite[\S~6.2]{esfahani2018data}. Unfortunately, problem~\eqref{eq:lipschitz-computation} maximizes a convex function over a convex set and is therefore hard. More formally, assume that $\ell(\xi)=\mu^\top\xi+\|\cov^{\frac{1}{2}} \xi\|_2$ for an arbitrary $\mu\in\RR^m$ and $\cov\in\PSD^m$. An elementary calculation shows that $\ell^*(z)=0$ if $z\in\mathcal E$ and $\ell^*(z)=\infty$ if $z\notin\mathcal E$, where $\mathcal E=\{\mu+\cov^{\frac{1}{2}}u:\|u\|_2\le 1\}$ stands for the ellipsoid with center $\mu$ and shape matrix $\cov$. Hence, $\mathcal E$ is the domain of $\ell^*(z)$. In order to compute the Lipschitz modulus of $\ell(\xi)$ with respect to the $\infty$-norm, for example, we thus need to solve an instance of problem~\eqref{eq:lipschitz-computation} that maximizes the 1-norm over $\mathcal E$. As maximizing the 1-norm over an arbitrary ellipsoid is NP-hard \cite[Lemma~4.1]{hanasusanto2015distributionally}, we conclude that the worst-case risk evaluation problem~\eqref{eq:worst:risk} is intractable even for polyhedral norms and for simple classes of (convex) conic quadratic loss functions. \hfill $\Box$
\end{remark}

One can show that the supremum of the worst-case risk evaluation problem~\eqref{eq:worst:risk} is {\em never} attained under the conditions of Theorem~\ref{thm:type1:convex}, that is, any asymptotically optimal sequence of distributions must push some (decreasing amount of) probability mass to infinity. As in the case of a piecewise concave loss function, such a sequence can be constructed explicitly. To do so, choose a maximizer $z\opt$ of problem~\eqref{eq:lipschitz-computation}, which is generally intractable as pointed out in Remark~\ref{rem:lipschitz}. Moreover, select $i_0 \in [N]$ and $\xi\opt\in\arg\max_{\| \xi \| \leq 1} \xi^\top z\opt$. Then, the distributions
    \begin{equation*}
        \QQ_n = \frac{1}{N} \sum_{i \neq i_0}^N \delta_{\widehat \xi_i} + \frac{n-1}{n N} \, \delta_{\widehat \xi_{i_0}}
        + \frac{1}{n N} \, \delta_{\widehat \xi_{i_0} + \eps n N \xi\opt}
    \end{equation*}
    can be shown to be feasible and asymptotically optimal in~\eqref{eq:worst:risk} as $n\ge 1$ tends to infinity.


Assume next that $p=2$, the loss function $\ell(\xi)$ is quadratic and the transportation cost in the definition of the Wasserstein distance is induced by the Euclidean norm. Then, the worst-case risk~\eqref{eq:worst:risk} coincides with the optimal value of a tractable semidefinite program~(SDP).

\begin{theorem}[Indefinite quadratic loss and $p=2$ I]
\label{theorem:convex:quadratic}
Assume that $\Xi = \RR^m$ and that $\ell(\xi) = \xi^\top Q \xi + 2 q^\top \xi$ with $Q \in \mathbb S^m$ and $q \in \RR^m$ is a (possibly indefinite) quadratic loss function. If $p = 2$, $\| \cdot\| = \| \cdot \|_2$ is the Euclidean norm on $\RR^m$ and $\Pnom_N$ is the empirical distribution, then the worst-case risk~\eqref{eq:worst:risk} coincides with the optimal value of a tractable SDP, that is,\footnote{Updated.}
\begin{equation}
\label{eq:soroosh-convex}
\optimize{\risk_{\eps, 2}(\Pnom_N, \ell) ~=~ \inf & \DS \gamma \eps^2 + \frac{1}{N} \sum_{i=1}^N s_i \\
\hspace{7em} \st & \gamma \in \RR_+, \; s_i \in \RR & \forall i \in [N] \\[1mm]
& \begin{bmatrix}
\gamma I - Q & q + \dualvar \widehat \xi_i \\ q^\top + \dualvar \widehat \xi_i^\top  & ~ s_i +\dualvar \| \widehat \xi_i \|_2^2
\end{bmatrix} \succeq 0 & \forall  i \in [N]\, .}
\end{equation}
\end{theorem}


Note that substituting the SDP~\eqref{eq:soroosh-convex} into the distributionally robust decision problem~\eqref{eq:dro} yields a tractable SDP if the set $\Lc$ of admissible loss functions is defined through SDP constraints in $Q$ and $q$. In order to construct an extremal distribution that solves problem~\eqref{eq:worst:risk} for a {\em fixed} convex quadratic loss function, it is useful to derive the dual of the SDP~\eqref{eq:soroosh-convex}.

\begin{theorem}[Indefinite quadratic loss and $p=2$ II]
\label{theorem:convex:quadratic:extremal}
Suppose that all conditions of Theorem~\ref{theorem:convex:quadratic} hold. If $\lambda_{\rm max}(Q)$ denotes the largest eigenvalue of $Q$, then
\begin{equation}
\label{eq:soroosh-convex-dual}
\optimize{\risk_{\eps, 2}(\Pnom_N, \ell) ~=~ \max & \DS \frac{1}{N} \sum_{i=1}^N (\widehat \xi_i+ \theta_i)^\top Q (\widehat \xi_i+ \theta_i)+2q^\top(\widehat \xi_i+ \theta_i) + \alpha \lambda_{\rm max}(Q) \\ [3ex]
\hspace{7em} \st & \alpha \in \RR_+, ~ \theta_i \in \RR^m \quad \forall i \in [N] \\ [1ex]
& \DS \frac{1}{N} \sum_{i=1}^N \| \theta_i \|_2^2 + \alpha \leq \eps^2. }
\end{equation}
\end{theorem}

Problem~\eqref{eq:soroosh-convex-dual} represents a quadratically constrained quadratic program (QCQP) with a compact feasible set and is therefore solvable. As $Q$ is not necessarily negative semidefinite, problem~\eqref{eq:soroosh-convex-dual} is generally {\em nonconvex}. This is perhaps puzzling because~\eqref{eq:soroosh-convex-dual} is obtained by `massaging' the dual of~\eqref{eq:soroosh-convex} and because dual optimization problems are convex by construction. The apparent contradiction is resolved by noting that nonconvex QCQPs of the form~\eqref{eq:soroosh-convex-dual} with a {\em single} constraint are equivalent to convex SDPs by virtue of the celebrated $\mathcal S$-procedure \cite[Appendix~B.1]{boyd2004convex}.

Intuitively, problem~\eqref{eq:soroosh-convex-dual} can be interpreted as a {\em finite reduction} of the worst-case risk evaluation problem~\eqref{eq:worst:risk}, which maximizes only over discrete distributions in the Wasserstein ball. Denoting by $v_{\rm max}(Q)$ an eigenvector corresponding to $\lambda_{\rm max}(Q)$, any such discrete distribution assigns probability $1/N$ to the perturbed training samples $\widehat \xi_i+ \theta_i$, $i\in[N]$, and a `vanishing' probability to an atom located `infinitely' far away in the direction of~$v_{\rm max}(Q)$. More precisely, the product of the squared transportation distance and the probability of this last atom must converge to a finite value~$\alpha\in\RR_+$ (hence, the probability of this atom must be asymptotically proportional to the inverse of the squared transportation distance). In the same spirit one can show that if $\alpha\opt$ and $\{\theta_i\opt\}_i$ are optimal in~\eqref{eq:soroosh-convex-dual} and $i_0\in[N]$, then the discrete distributions
\begin{equation*}
\QQ_n = \frac{1}{N} \sum_{i \neq i_0}^N \delta_{\widehat\xi_i+\theta_i\opt} + \frac{n-1}{n N} \, \delta_{\widehat\xi_{i_0}+\theta_{i_0}\opt} + \frac{1}{n N} \, \delta_{\widehat \xi_{i_0} + \sqrt{n N \alpha\opt} \; v_{\rm max}(Q)}
\end{equation*}
are feasible and asymptotically optimal in~\eqref{eq:worst:risk} as $n\ge 1$ tends to infinity.

The structure of the extremal distributions for the worst-case risk evaluation problem~\eqref{eq:worst:risk} with general loss functions and nominal distributions as well as necessary and sufficient conditions for their existence have been studied in~\cite{gao2016distributionally, owhadi2017extreme, wozabal2012framework}. The special case of a Wasserstein ball centered at a discrete distribution with $N$ atoms has undergone particular scrutiny. 
Considerable effort was spent on proving the existence of discrete extremal distributions with as few atoms as possible. A first breakthrough was marked by the insight that the worst-case risk of any continuous bounded loss function is attained by a discrete distribution with at most $N+3$ atoms \cite[Theorem~2.3]{wozabal2012framework}. As any $(N+3)$-point distribution on $\RR^m$ can be encoded by $(N+3)\cdot(m + 1)-1$ parameters ({\em i.e.}, the coordinates and probabilities of the $N+3$ atoms), this result motivates a {\em finite reduction}: when searching for an extremal distribution, one may restrict attention to discrete distributions supported on $N+3$ points, which amounts to searching a finite-dimensional parameter space. It was later shown that one may actually focus on discrete distributions with $N+2$ atoms~\cite[Theorem~2.3]{owhadi2017extreme} or even only $N+1$ atoms~\cite[Corollary~1]{gao2016distributionally} without sacrificing optimality. These sharper results facilitate more parsimonious finite reductions that may be fruitfully used in algorithm design. Exact finite reductions involving fewer atoms are available only in special cases. For example, the discussion after Theorem~\ref{thm:type1:extrem} shows that the worst-case risk of a {\em concave} loss function is always attained by an $N$-point distribution. For more general loss functions, however, every $N$-point distribution may be suboptimal even if the worst-case risk is attained. 

\begin{example}[Non-existence of extremal distributions with $N$ atoms]
\label{ex:non-existence2}
~ Suppose that $\Xi = (-\infty, 2] $, $N = 1$ and $\widehat \xi_1 = 0$, which implies that $\Pnom_1$ is the Dirac distribution at~$0$. Set the norm on $\RR$ to the absolute value $|\cdot|$, select $\eps\in(0,2)$ and set $\ell(\xi ) = \max \{ 0, \xi-1 \}$. Next, define $\QQ\opt = (1 - \eps / 2 ) \, \delta_{0} + ( \eps / 2 ) \, \delta_{2} $, and note that the type-1 Wasserstein distance between $\Pnom_1$ and $\QQ\opt$ amounts to $\eps$, which is the cost of moving mass $\eps/2$ from 2 to 0. Thus, $\QQ\opt\in \B_{\eps,1}(\Pnom_1)$. Moreover, we have $\EE^{\QQ\opt}[\ell(\xi)]= \eps/2$, which provides a lower bound on the worst-case risk. In fact, by solving problem~\eqref{dual-dual} one can show that $\QQ\opt$ is optimal in~\eqref{eq:worst:risk}. Any one-point distribution $\delta_z$ resides in the Wasserstein ball of radius $\eps$ only if $|z|\le \eps$, and therefore the maximum risk that any one-point distribution can attain is $\max\{0, \eps-1\}$, which is strictly smaller than $\eps/2$ for any $\eps \in (0,2)$. Thus, no one-point distribution can be extremal. \hfill $\Box$
\end{example}

If the worst-case risk over a Wasserstein ball centered at the empirical distribution is attained, then there always exists an extremal distribution with $N+1$ atoms that can be characterized in quasi-closed form~\cite[Corollary~2]{gao2016distributionally}. In practice, however, it is often convenient to ignore this minimal representability and to search over candidate distributions with more than $N+1$ atoms, {\em e.g.}, by solving a finite convex optimization problem such as~\eqref{thm:type1:extrem}. For generic nominal distributions, necessary and sufficient conditions for the existence of an extremal distribution are detailed in~\cite[Corollary~1]{gao2016distributionally}.

\subsection{Tractability Results for Elliptical Nominal Distributions}
\label{sec:wc-risk-elliptical}

We will now demonstrate that the worst-case risk evaluation problem~\eqref{eq:worst:risk} and the distributionally robust decision problem~\eqref{eq:dro} sometimes admit exact tractable reformulations or conservative tractable approximations even if the nominal distribution $\Pnom$ is continuous. To show this, we assume throughout this section that $\Pnom_N$ has mean vector $\msa\in\RR^m$ and covariance matrix $\covsa\in \PSD^m$. Thus, we implicitly assume that~$\Pnom_N$ has finite second-order moments.


We first define an uncertainty set in the space of mean vectors and covariance matrices.
\[
\mathcal U_{\eps}(\msa, \covsa) = 
\left\{ (\m,\cov) \in \RR^m\times \PSD^m: \| \msa - \m \|_2^2 + \Tr{\covsa + \cov - 2 \left( \covsa^{\frac{1}{2}} \cov \covsa^{\frac{1}{2}} \right)^{\frac{1}{2}} } \leq \eps^2 \right\}
\]

This uncertainty set is of interest because it covers the projection of the type-2 Wasserstein ball $\B_{\eps,2}(\Pnom_N)$ onto the space of mean vectors and covariance matrices. Moreover, if the nominal distribution is elliptical, $\mathcal U_{\eps}(\msa, \covsa)$ is actually equal to the projection of $\B_{\eps,2}(\Pnom_N)$.

\begin{proposition}[Projection of $\B_{\eps,2}(\Pnom_N)$ onto the mean-covariance space]
\label{prop:gelbrich-projection} 
If $\Pnom_N$ has mean vector $\msa\in\RR^m$ and covariance matrix $\covsa\in \PSD^m$, then
\[
     \left\{ \left(\EE^{\QQ}[\xi] , \EE^\QQ[(\xi-\EE^{\QQ}[\xi])(\xi-\EE^{\QQ}[\xi])^\top] \right) : \QQ\in \B_{\eps,2}(\Pnom_N)\right\} \subseteq \mathcal U_{\eps}(\msa, \covsa).
\]
The inclusion becomes an equality if $\Xi=\RR^m$ and $\Pnom_N= \mathcal{E}_g(\widehat \mu, \widehat \Sigma)$ is an elliptical distribution.
\end{proposition}

Proposition~\ref{prop:gelbrich-projection} follows immediately from Theorem~\ref{theorem:gelbrich}. The condition $\Xi=\RR^m$ ensures that any elliptical distribution $\QQ= \mathcal{E}_g(\mu, \Sigma)$ with the same density generator as the nominal distribution and with $W_2(\QQ,\Pnom_N)\le\eps$ belongs to $\B_{\eps,2}(\Pnom_N)$. One can show that~$\mathcal U_{\eps}(\msa, \covsa) $ is convex and compact~\cite[Lemma~A.6]{shafieezadeh2018wasserstein}, which is expected as it is a projection of a (Wasserstein)~ball.

The uncertainty set $\mathcal U_{\eps}(\msa, \covsa)$ can conveniently be used in classical robust optimization. Indeed, a robust constraint that requires a concave function $h(\mu,\cov)$ to be nonpositive for all $(\mu,\cov)\in \mathcal U_{\eps}(\msa, \covsa)$ can be reformulated as a convex constraint that involves the conjugate of $-h(\mu,\cov)$ and the support function of the uncertainty set $\mathcal U_{\eps}(\msa, \covsa)$ \cite[Theorem~2]{ben2015deriving}, that~is,
\[
    h(\mu,\cov) \le 0\quad \forall (\mu,\cov)\in \mathcal U_{\eps}(\msa, \covsa)\quad \iff\quad\left\{ \begin{array}{l} \exists q\in\RR^m,\,Q\in\mathbb S^m:\\[1ex]
    (-h)^*(-q,-Q)+\sigma_{\mathcal U_{\eps}(\msa, \covsa)}(q, Q)\le 0.\end{array}\right.
\]
This constraint is computationally tractable for many commonly used constraint functions because the support function of $\mathcal U_{\eps}(\msa, \covsa)$ is SDP-representable~\cite{nguyen2019distributionally}.

\begin{lemma}[Support function of $\mathcal U_\eps(\msa, \covsa)$]
    The support function of $\mathcal U_{\eps}(\msa, \covsa)$ coincides with the optimal value of a tractable SDP, that is, for any $q\in\RR^m$ and $Q\in\mathbb S^m$ we~have\footnote{Updated.}
    \begin{align*}
    \begin{array}{cc@{\quad}l}
    \sigma_{\mathcal U_{\eps}(\msa, \covsa)}(q, Q)\; =
    & \inf & q^\top \msa + \tau + \dualvar \left(\eps^2 - \mathop{\rm Tr}\, [\covsa]\right) + \Tr{Z} \\[1ex]
    & \st & \gamma \in \RR_+, \; \tau \in\RR_+,\; Z \in \mathbb{S}^m_+ \\[1ex]
    & & \begin{bmatrix} \gamma I - Q & \gamma \covsa^{\frac{1}{2}} \\ \gamma \covsa^{\frac{1}{2}} & Z \end{bmatrix} \succeq 0, \quad \left\| \begin{pmatrix} q\\ \tau-\gamma \end{pmatrix} \right\|_2\le \tau+\gamma.
    \end{array}
    \end{align*}
\end{lemma}

Unlike the mean vector $\mu= \EE^{\QQ}[\xi]$ and the second-order moment matrix $M=\EE^{\QQ}[\xi\xi^\top]$, both of which constitute linear functions of the underlying distribution $\QQ$, the covariance matrix $\cov=M-\mu\mu^\top$ is nonlinear in $\QQ$. The condition $(\mu,\cov)\in \mathcal U_\eps(\msa, \covsa)$ thus appears to be nonconvex in $\QQ$. To gain a clearer understanding, it is instructive to introduce the uncertainty set~$\mathcal V_{\eps}(\msa, \covsa)$ for~$(\mu,M)$ induced by the uncertainty set~$\mathcal U_{\eps}(\msa, \covsa)$ for~$(\mu,\cov)$, that is,
\[
\mathcal V_{\eps}(\msa, \covsa) = 
\left\{ (\m,M) \in \RR^m\times \PSD^m:  (\mu, M-\mu\mu^\top) \in \mathcal U_{\eps}(\msa, \covsa) \right\}.
\]
Maybe surprisingly, even though it is defined as the pre-image of a convex set under a {\em non}linear transformation, one can prove that $\mathcal V_{\eps}(\msa, \covsa)$ is convex. This implies, counterintuitively, that the condition $(\mu,\cov)\in \mathcal U_\eps(\msa, \covsa)$ is actually convex in $\QQ$ because it is equivalent to the requirement $(\mu,M)\in \mathcal V_\eps(\msa, \covsa)$ and because the moments $(\mu,M)$ are linear in $\QQ$.

Thanks to its convexity, the uncertainty set $\mathcal V_{\eps}(\msa, \covsa)$ can again conveniently be used in classical robust optimization. Indeed, a robust constraint that requires a concave function $h(\mu,M)$ to be nonpositive for all $(\mu,M)\in \mathcal V_{\eps}(\msa, \covsa)$ can be reformulated as a simple convex constraint involving the conjugate of $-h(\mu,M)$ and the support function of $\mathcal V_{\eps}(\msa, \covsa)$. This constraint is computationally tractable for many commonly used constraint functions because the support function of $\mathcal V_{\eps}(\msa, \covsa)$ is SDP-representable~\cite{nguyen2019distributionally}.

%
\begin{lemma}[Support function of $\mathcal V_\eps(\msa, \covsa)$]
    \label{lemma:V-support-fct}
    The support function of $\mathcal V_\eps(\msa, \covsa)$ coincides with the optimal value of a tractable SDP, that is, for any $q \in \RR^m$ and $Q \in \mathbb S^m$, we~have
    \begin{align} \label{eq:random}
    \begin{array}{cc@{\quad}l}
    \sigma_{\mathcal V_{\eps}(\msa, \covsa)}(q, Q) \; = 
    &\inf & \dualvar (\eps^2 - \|\msa\|_2^2 - \mathop{\rm Tr}[\covsa]) + z + \Tr{Z} \\
    &\st & \dualvar \in \RR_+, \, z \in \RR_+, \, Z \in \PSD^m \\[1ex]
    && \begin{bmatrix} \dualvar I - Q & \dualvar \covsa^{\frac{1}{2}} \\ \dualvar \covsa^{\frac{1}{2}} & Z \end{bmatrix} \succeq 0, ~ \begin{bmatrix} \dualvar I - Q & \dualvar \msa + \frac{q}{2} \\ (\dualvar \msa + \frac{q}{2})^\top & z \end{bmatrix} \succeq 0.
    \end{array}
    \end{align}
\end{lemma}

A useful ambiguity set in the space of probability distributions is the {\em Gelbrich hull}, which is constructed as the pre-image of $\mathcal U_{\eps}(\msa, \covsa)$ under the mean-covariance projection.

\begin{definition}[Gelbrich hull]
\label{def:gelbrich-hull}
The Gelbrich hull is given by 
\begin{equation*}
\label{eq:Gelbrich:ball}
\mathbb{G}_{\eps}(\msa, \covsa) = \left\{ \QQ \in \mathcal P(\Xi): 
\left(\EE^{\QQ}[\xi] , \EE^\QQ[(\xi-\EE^{\QQ}[\xi])(\xi-\EE^{\QQ}[\xi])^\top] \right)  \in \mathcal U_{\eps}(\msa, \covsa)
\right\}.
\end{equation*}
\end{definition}

By definition, $\mathbb{G}_{\eps}(\msa, \covsa)$ contains all distributions supported on $\Xi$ whose mean vectors and covariance matrices fall into the uncertainty set $\mathcal U_{\eps}(\msa, \covsa)$. Equivalently, by the definition of the induced uncertainty set $\mathcal V_\eps(\msa, \covsa)$, the Gelbrich hull can also be represented as
\[
    \mathbb{G}_{\eps}(\msa, \covsa) = \left\{ \QQ \in \mathcal P(\Xi):  \left(\EE^{\QQ}[\xi] , \EE^\QQ[\xi\xi^\top] \right)  \in \mathcal V_{\eps}(\msa, \covsa) \right\}.
\]
Thus, the Gelbrich hull can be expressed as the pre-image of the convex set $\mathcal V_{\eps}(\msa, \covsa)$ under a linear transformation, which shows that is is actually convex. We emphasize that convexity is not apparent from Definition~\ref{def:gelbrich-hull}, which introduces the Gelbrich hull as the pre-image of a convex set under a {\em non}linear transformation.

If we define $ \mathcal P(\Xi,\mu,\Sigma)$ as the {\em Chebyshev ambiguity set} that contains all distributions on~$\Xi$ with mean vector~$\mu$ and covariance matrix~$\cov$, then the Gelbrich hull can also be expressed~as
\begin{equation}
    \label{eq:gelbrich-union}
    \mathbb{G}_{\eps}(\msa, \covsa) = \bigcup_{(\mu,\Sigma) \in \mathcal U_{\eps}(\msa, \covsa)} \mathcal P(\Xi,\mu,\Sigma).
\end{equation}
From this representation it is evident that if $\mathbb{G}_{\eps}(\msa, \covsa)$ contains a distribution $\QQ$, then it contains {\em all} distributions on $\Xi$ that have the same mean vector and covariance matrix as $\QQ$. It is easy to verify that the Gelbrich hull provides an outer approximation for any Wasserstein ball $\B_{\eps,p}(\Pnom_N)$ with $p\ge 2$. Indeed, if $\B_{\eps,p}(\Pnom_N)$ contains a distribution $\QQ$ with mean vector~$\mu$ and covariance matrix~$\cov$, then $(\mu,\cov)\in\mathcal U_{\eps}(\msa, \covsa)$ by virtue of Proposition~\ref{prop:gelbrich-projection}, which implies via~\eqref{eq:gelbrich-union} that $\QQ\in \mathbb{G}_{\eps}(\msa, \covsa)$. These insights culminate in the following theorem.

\begin{theorem}[Gelbrich hull]
    \label{theorem:gelbrich:amb}
    If the nominal distribution $\Pnom_N$ has mean vector $\msa \in \RR^m$ and covariance matrix $\covsa \in \PSD^m$, then we have $\B_{\eps,p}(\Pnom_N) \subseteq \mathbb{G}_{\eps}(\msa, \covsa)$ for every $p \geq 2$.
\end{theorem}

Theorem~\ref{theorem:gelbrich:amb} shows that the Gelbrich hull provides an outer approximation for all Wasserstein balls $\B_{\eps,p}(\Pnom_N)$ with $p\ge 2$ solely on the basis of mean and covariance information. Discarding all information about $\Pnom_N$ beyond its first- and second-order moments can be seen as a compression of the training dataset. This amounts to sacrificing higher-order moment information and may improve the tractability of the risk evaluation problem~\eqref{eq:worst:risk} and the distributionally robust decision problem~\eqref{eq:dro}. To show this, we define the {\em Gelbrich risk} as
\begin{equation}
\label{eq:gelbrich:risk}
\overline \risk_{\eps}(\msa, \covsa, \ell) = \sup_{\QQ \in \mathbb{G}_{\eps}(\msa, \covsa)} ~ \risk(\QQ, \ell)
\end{equation}
and the {\em optimal Gelbrich risk} as
\begin{equation}
\label{eq:gelbrich-dro}
\overline\risk_{\eps}(\msa, \covsa, \Lc) = \inf_{\ell \in \Lc} ~ \overline \risk_{\eps}(\msa, \covsa, \ell).
\end{equation}
Theorem~\ref{theorem:gelbrich:amb} immediately implies that the (optimal) Gelbrich risk provides an upper bound on the (optimal) worst-case risk whenever $p\ge 2$. 
\begin{corollary}[Gelbrich risk]
\label{cor:gelbrich-risk}
If the nominal distribution $\Pnom_N$ has mean vector $\msa \in \RR^m$ and covariance matrix $\covsa \in \PSD^m$ and if $p\ge 2$, then 
\[
    \risk_{\eps, p}(\Pnom_N, \ell)\le \overline \risk_{\eps}(\msa, \covsa, \ell)\quad \forall \ell\in\Lc \qquad \text{and}\qquad \risk_{\eps, p}(\Pnom_N, \Lc) \le \overline\risk_{\eps}(\msa, \covsa, \Lc).
\]
\end{corollary}

The representation~\eqref{eq:gelbrich-union} of the Gelbrich hull as a union of Chebyshev ambiguity sets suggests that the Gelbrich risk of any fixed loss function~$\ell(\xi)$ can be expressed as the optimal value of the following two-layer optimization problem~\cite{nguyen2019distributionally}.
    \begin{subequations}
    \begin{align}
    \label{eq:two-layer-a}
    \overline \risk_{\eps}(\msa, \covsa, \ell)~& = \Sup_{(\mu, \cov) \in \mathcal U_\eps(\msa, \covsa)} ~\,\quad \Sup_{\QQ \in \mathcal P(\Xi,\mu,\Sigma)} \risk(\QQ, \ell)\\
    & = \Sup_{(\mu, M) \in \mathcal V_\eps(\msa, \covsa)} \, \Sup_{\QQ \in \mathcal P(\Xi,\mu,M-\mu\mu^\top)} \risk(\QQ, \ell) \label{eq:two-layer-b}
    \end{align}
    \end{subequations}
Note that~\eqref{eq:two-layer-b} follows immediately from the definition of the uncertainty set $\mathcal V_\eps(\msa, \covsa)$ and the formula for the covariance matrix in terms of the mean vector and the second-order moment matrix. The inner problems in~\eqref{eq:two-layer-a} and~\eqref{eq:two-layer-b} both represent the same distributionally robust optimization problem over a Chebyshev ambiguity set but with different parameterizations. This problem can be viewed as an infinite-dimensional linear program over all probability distributions $\QQ$ that satisfy the linear equality constraints $\EE^\QQ[\xi]=\mu$ and $\EE^\QQ[\xi\xi^\top]=M$. Therefore, the optimal value of the inner maximization problem is concave in the right hand side parameters~$\mu$ and~$M$ but generally nonconcave in the alternative parameters $\mu$ and $\cov$. The outer problem in~\eqref{eq:two-layer-a} hedges against ambiguity in the mean vector and the covariance matrix, while the one in~\eqref{eq:two-layer-b} hedges against ambiguity in the first- and second-order moments. The formulation~\eqref{eq:two-layer-a} is conceptually appealing because of its connection to the Wasserstein distance and because it is more natural to characterize a distribution in terms of its mean vector and covariance matrix. The formulation~\eqref{eq:two-layer-b}, on the other hand, is computationally attractive because it expresses the outer problem as a convex program that maximizes a manifestly concave function over the convex set $\mathcal V_\eps(\msa, \covsa)$.

\begin{remark}[Second layer of robustness]
\label{rem:2nd-layer-robustness}
Distributionally robust optimization problems akin to~\eqref{eq:two-layer-a} and~\eqref{eq:two-layer-b} that accommodate a second layer of robustness to account for moment ambiguity have been investigated in \cite{delage2010distributionally, el-ghaoui2003wcvar, hanasusanto2015distributionally, natarajan2010utility, rujeerapaiboon2015robust, zymler2013wcvar}, among others. As the optimal value of the inner maximization problem is always concave in~$(\mu, M)$ but typically nonconcave in~$(\mu,\cov)$, moment ambiguity has mostly been modeled through convex uncertainty sets for~$(\mu,M)$, thereby ensuring convexity of the outer maximization problem. For example, uncertainty sets that force $\mu$ to lie in an ellipsoid and $M$ in the intersection of two positive semi-definite cones were studied in \cite{delage2010distributionally}, while box-type uncertainty sets for~$(\mu,M)$ were proposed in~\cite{natarajan2010utility} and refined in~\cite{hanasusanto2015distributionally, zymler2013wcvar}. Convex uncertainty sets for $(\mu,\cov)$ were shown to render the outer maximization problems convex only in special cases, {\em e.g.}, when evaluating a worst-case value-at-risk of a linear or quadratic loss function~\cite{el-ghaoui2003wcvar, rujeerapaiboon2015robust}. The convex uncertainty set $\mathcal U_\eps(\msa, \covsa)$ for $(\mu,\cov)$ is remarkable because it leads to a second-layer maximization problem in~\eqref{eq:two-layer-a} that admits a convex reformulation for {\em all} loss functions~$\ell(\xi)$.~\hfill$\Box$
\end{remark}

The decomposition~\eqref{eq:two-layer-b} of the Gelbrich risk evaluation problem into two consecutive maximization problems offers a systematic approach to derive convex reformulations for~\eqref{eq:gelbrich:risk}. A tractable SDP reformulation is available, for example, when the loss function $\ell(\xi)$ is a pointwise maximum of finitely many (possibly indefinite) quadratic functions.

%

\begin{theorem}[Piecewise quadratic loss I]
    \label{theorem:p.w.quadratic}
    Assume that $\Xi = \RR^m$ and $\eps>0$ and that $\ell(\xi) = \max_{j \in [J]} \{\xi^\top Q_j \xi + 2 q_j^\top \xi + q^0_{j} \}$ with $Q_j \in \mathbb S^m$, $q_j \in \RR^m$, and $q_{j}^0 \in \RR$ for any $j \in [J]$ is a piecewise quadratic loss function. If $\msa\in \RR^m$ and $\covsa \in \PSD^m$, then the Gelbrich risk~\eqref{eq:gelbrich:risk} is equal to the optimal value of a tractable SDP, that is,
    \begin{equation}\label{eq:p.w.quad} 
    \begin{array}{rcl}
    \overline \risk_{\eps}(\msa, \covsa, \ell) ~=&\inf & y_0 + \dualvar \left(\eps^2 - \| \msa \|_2^2 -  \mathop{\rm Tr}[\covsa]\right) + z + \Tr{Z} \\[2ex]
    &\st &\dualvar \in \RR_+, \; y_0 \in \RR, \; y \in \RR^m, \; Y \in \mathbb{S}^m, \; z \in \RR_+, \; Z \in \PSD^m \\[1ex]
    && \begin{bmatrix} \dualvar I - Y & y + \dualvar \msa \\ y^\top + \dualvar \msa^\top & z \end{bmatrix} \succeq 0, \quad \begin{bmatrix} \dualvar I - Y & \dualvar \covsa^{\frac{1}{2}} \\ \dualvar \covsa^{\frac{1}{2}} & Z \end{bmatrix} \succeq 0\\
    &&\begin{bmatrix} Y - Q_j & y - q_j \\[0.5ex] y^\top - q_j^\top & y_0 - q^0_{j} \end{bmatrix} \succeq 0 \quad \forall j \in [J].
    \end{array}
    \end{equation}
\end{theorem}

In order to construct an extremal distribution for the Gelbrich risk evaluation problem~\eqref{eq:gelbrich:risk}, it is again expedient to derive the dual of the SDP~\eqref{eq:p.w.quad}. 

\begin{theorem}[Piecewise quadratic loss II]
    \label{thm:II}
    If all conditions of Theorem~\ref{theorem:p.w.quadratic} hold, then we have
    \begin{equation}
    \label{theorem:p.w.quadratic2}
    \optimize{
    \overline \risk_{\eps}(\widehat \mu, \widehat \Sigma, \ell) =  \max & \DS \sum_{j=1}^J \Tr{Q_j \Theta_j} + 2 q_j^\top \theta_j + q^0_{j} \alpha_j \\ [3ex]
    \hspace{5.5em}\st & \DS \mu\in \RR^m, \;  \Sigma\in \PSD^m,\; \alpha_j\in\RR_+, \;  \theta_j \in \RR^m,\; \Theta_j \in \PSD^m \quad \forall j \in [J] \\ [1ex]
    & \begin{bmatrix}
    \Theta_j & ~\theta_j \\[1ex] \theta_j^\top & ~\alpha_j
    \end{bmatrix} \succeq 0 \quad \forall j \in [J] \\ [3ex]
    & \DS \sum_{j=1}^J \alpha_{j} = 1 ,~  \DS \sum_{j=1}^J \theta_j = \mu ,~  \DS \sum_{j=1}^J \Theta_{j} = \Sigma+ \mu \mu^\top, ~ (\mu,\cov)\in\mathcal U_\eps(\msa, \covsa).}
    \end{equation}
\end{theorem}

Note that problem~\eqref{theorem:p.w.quadratic2} has a continuous objective function as well as a compact feasible set and is therefore solvable. Any optimal solution $(\mu\opt,\cov\opt, \{\alpha_j\opt, \theta_j\opt, \Theta_j\opt \}_j)$ can in principle be used to construct an extremal distribution~$\QQ\opt$ that attains the supremum in the Gelbrich risk evaluation problem~\eqref{eq:gelbrich:risk}.\footnote{Updated.} Define the index sets
\begin{align*}
    \nu_+ = \{ j \in [J] : \alpha\opt_j > 0 \} \quad
    \text{and} \quad \nu_\infty = \{ j \in [J] : \alpha\opt_j = 0, ~ \Theta\opt_j \neq 0 \}.
\end{align*}
If $\nu_\infty = \emptyset$, it is easy to show that a mixture of Gaussian distributions is optimal in~\eqref{eq:gelbrich:risk}. Specifically, define the distribution $\QQ\opt_j = \mathcal N(\theta\opt_j / \alpha\opt_j, \Theta\opt_j / \alpha\opt_j)$ for every $j \in \nu_+$. We next show that the mixture distribution $\QQ\opt = \sum_{j\in \nu_+} \alpha\opt_j \, \QQ\opt_j$ is optimal in~\eqref{eq:gelbrich:risk}. By construction, $\QQ\opt$ has mean vector $\mu\opt$ and covariance matrix $\cov\opt$. Thus, we may conclude that $\QQ^\star \in \mathbb G_\eps(\msa, \covsa)$. Furthermore, the definition of~$\QQ^\star$ as a mixture distribution and the definition of~$\ell$ as a pointwise maximum of quadratic component functions implies that
\begin{align*}
    \textstyle
    \EE^{\QQ^\star}[\ell(\xi)] 
    \geq \sum_{j \in \nu_+} \alpha_j^\star \, \EE^{\QQ_j^\star}[\xi^\top Q_j \xi + 2 q_j^\top \xi + q_j^0]  
    = \sum_{j \in [J]} \Tr{Q_j \Theta_j^\star} + 2 q_j^\top \theta_j^\star + q_j^0 p_j^\star.
\end{align*} 
Specifically, the inequality holds because $\ell(\xi) \geq \xi^\top Q_j \xi + 2 q_j^\top \xi + q_j^0$ for every~$j\in[J]$, and the equality holds because $\nu_\infty = \emptyset$ and $\theta^\star_j = 0$ whenever $\alpha_j^\star = 0$.  This ensures that $\QQ^\star$ solves the worst-case expectation problem~\eqref{eq:gelbrich:risk}.
If $\nu^\infty \neq \emptyset$, one can also show that the distributions
\begin{equation*}
    \QQ_n = \DS \sum_{j\in\nu_+\cup \nu_\infty} \alpha_{j}(n)  \, \mathcal N(\theta\opt_j / \alpha_j(n), \Theta\opt_j / \alpha_j(n))
    \quad \text{with} \quad
    \alpha_j(n) = \begin{cases}
        \alpha\opt_j \left( 1 - \tfrac{|\nu_\infty|}{n} \right) & \text{if}~j \in \nu_+ \\
        \frac{1}{n} & \text{if}~j \in \nu_\infty
    \end{cases}
\end{equation*}        
are feasible and asymptotically optimal in~\eqref{eq:worst:risk} as $n\ge |\nu_\infty|$ tends to infinity.

While exactly computable in polynomial time, the Gelbrich risk of a piecewise quadratic loss function may only provide a loose upper bound on the worst-case risk under the Wasserstein ambiguity set, which is often the actual quantity of interest. One can prove, however, that the Gelbrich risk~\eqref{eq:gelbrich:risk} coincides with the worst-case risk~\eqref{eq:worst:risk} with respect to a type-2 Wasserstein ball if the loss function is quadratic and the nominal distribution is elliptical.

\begin{theorem}[Indefinite quadratic loss I]
\label{thm:single:quad}
Assume that $\Xi = \RR^m$ and that $\ell(\xi) = \xi^\top Q \xi + 2 q^\top \xi$ with $Q \in \mathbb S^m$ and $q \in \RR^m$ is a quadratic loss function. If $\msa\in \RR^m$ and $\covsa\in\PSD^m$, then the Gelbrich risk~\eqref{eq:gelbrich:risk} is equal to the optimal value of a tractable SDP, that is,
\begin{equation}\label{eq:gelbrich:quad}
\begin{array}{rcl}
\overline \risk_{\eps}(\msa, \covsa, \ell) ~=&\inf & \dualvar \left(\eps^2 - \|\msa\|_2^2 - \mathop{\rm Tr}\, [\covsa]\right) + z  + \Tr{Z} \\[2ex]
&\st &\dualvar \in \RR_+, \; z \in \RR_+,\; Z \in \PSD^m \\[1ex]
&&\begin{bmatrix} \dualvar I - Q & q + \dualvar \msa \\ q^\top + \dualvar \msa^\top & z \end{bmatrix} \succeq 0 ,\quad \begin{bmatrix} \dualvar I - Q & \dualvar \covsa^{\frac{1}{2}} \\ \dualvar \covsa^{\frac{1}{2}} & Z \end{bmatrix} \succeq 0.
\end{array}
\end{equation}
Moreover, if $\Pnom_N= \mathcal{E}_g(\widehat \mu, \widehat \Sigma)$ is an elliptical distribution with mean vector $\msa$ and covariance matrix $\covsa$, $p = 2$ and $\| \cdot\| = \| \cdot \|_2$ is the Euclidean norm on $\RR^m$, then the worst-case risk~\eqref{eq:worst:risk}, the Gelbrich risk~\eqref{eq:gelbrich:risk} and the optimal value of the SDP~\eqref{eq:gelbrich:quad} are all equal. 
\end{theorem}

%

The SDP~\eqref{eq:gelbrich:quad} is easily obtained from~\eqref{eq:p.w.quad} by setting $J=1$ and noting that $Y=Q_1$, $y=q_1$ and $y_0=0$ at optimality. As usual, a discrete extremal distribution $\QQ\opt$ for the Gelbrich risk evaluation problem~\eqref{eq:gelbrich:risk} can be derived from the dual of the SDP~\eqref{eq:gelbrich:quad}. In the following we denote the mean vector and the covariance matrix of $\QQ\opt$ by~$\mu\opt$ and~$\cov\opt$, respectively. As $\QQ\opt\in\mathbb{G}_{\eps}(\msa, \covsa)$, and as the Gelbrich hull is constructed solely on the basis of first- and second-order moment information, {\em any} distribution with mean vector $\mu\opt$ and covariance matrix $\cov\opt$ belongs to $\mathbb{G}_{\eps}(\msa, \covsa)$, too. Moreover, as $\ell(\xi)$ is quadratic, the risk $\risk(\QQ\opt,\ell)$ depends on $\QQ\opt$ only through its first- and second-order moments. This implies that {\em any} distribution with mean vector~$\mu\opt$ and covariance matrix $\cov\opt$ is optimal in~\eqref{eq:gelbrich:risk}. 

Consider now the problem of evaluating the worst-case risk~\eqref{eq:worst:risk} of the quadratic loss function $\ell(\xi)$ over a type-2 Wasserstein ball centered at an elliptical nominal distribution $\Pnom_N = \mathcal{E}_g(\widehat \mu, \widehat \Sigma)$. Theorem~\ref{theorem:gelbrich} ensures that all elliptical distributions in the Gelbrich hull with the same density generator as $\Pnom_N$ belong to the Wasserstein ball $\B_{\eps,2}(\Pnom_N)$. This implies that the special elliptical distribution $\QQ\opt = \mathcal E_g(\mu\opt, \cov\opt)$ is feasible in~\eqref{eq:worst:risk}. Moreover, we have
\[
    \overline \risk_{\eps}(\msa, \covsa, \ell) = \risk(\QQ\opt,\ell)\le \risk_{\eps, p}(\Pnom_N, \ell)\le \overline \risk_{\eps}(\msa, \covsa, \ell),
\]
where the equality holds because $\QQ\opt$ is optimal in the Gelbrich risk evaluation problem~\eqref{eq:gelbrich:risk}, while the two inequalities follow from the feasibility of $\QQ\opt$ in the worst-case risk evaluation problem~\eqref{eq:worst:risk} and Corollary~\ref{cor:gelbrich-risk}, respectively. Thus, all inequalities in the above expression are exact, which implies that $\QQ\opt$ is actually optimal in~\eqref{eq:worst:risk}.


Next, we show how $\QQ\opt$ can be constructed from the optimality conditions of the SDP~\eqref{eq:gelbrich:quad}.

\begin{theorem}[Indefinite quadratic loss II]
\label{thm:extremal:easy}
If all conditions of Theorem~\ref{thm:single:quad} hold, $\covsa \succ 0$ and there exists $\dualvar\opt\ge 0$ with $\dualvar\opt I \succ Q$ that solves the nonlinear algebraic equation
\begin{align}
     \| \widehat \mu - (\dualvar I -Q)^{-1} (q + \dualvar \widehat \mu) \|_2^2 + \Tr{\covsa \left( I - \dualvar (\dualvar I - Q)^{-1} \right)^2} =\eps^2,
    \label{eq:FOC:1}
\end{align}
then the Gelbrich risk~\eqref{eq:gelbrich:risk} is attained by any distribution with mean vector
\begin{subequations}
\begin{align}
\m\opt = ( \dualvar\opt I - Q)^{-1} (\dualvar\opt \widehat \mu + q) \label{eq:extremal:m}
\end{align}
and covariance matrix
\begin{align}
\cov\opt = (\dualvar\opt)^2 (\dualvar\opt I - Q)^{-1} \covsa (\dualvar\opt I - Q)^{-1}. \label{eq:extremal:cov}
\end{align}
\end{subequations}
Moreover, if $\Pnom_N = \mathcal{E}_g(\widehat \mu, \widehat \Sigma)$ is elliptical, $p = 2$ and $\| \cdot\| = \| \cdot \|_2$ is the Euclidean norm, then the elliptical distribution $\QQ\opt = \mathcal E_g(\mu\opt, \cov\opt)$ attains the worst-case risk in~\eqref{eq:worst:risk}.
\end{theorem}

One can show that if $Q\succeq 0$, then $\dualvar\opt$ exists and $\cov\opt \succeq \lambda_{\min}(\covsa) I$. To give an intuition for Theorem~\ref{thm:extremal:easy}, note that the SDP~\eqref{eq:gelbrich:quad} can be converted to an equivalent nonlinear program (NLP) in the single decision variable $\gamma$ by using Schur complements to show that
\[
    z=(q + \dualvar \msa)^\top(\dualvar I-Q)^{-1} (q + \dualvar \msa)\quad \text{and} \quad Z= \dualvar^2\; \covsa^{\frac{1}{2}} (\dualvar I-Q)^{-1} \covsa^{\frac{1}{2}}
\]
at optimality. The resulting NLP minimizes a strictly convex objective function that explodes as~$\dualvar$ drops to $ \lambda_{\max}(Q)$ or as $\dualvar$ tends to infinity. Equation~\eqref{eq:FOC:1} represents its first-order optimality condition, whose unique solution $\dualvar\opt$ can be computed efficiently to any precision via bisection or the Newton-Raphson method. 
Using~\eqref{eq:FOC:1}, one can then show that any distribution with mean vector $\mu\opt$ and covariance matrix $\cov\opt$ as defined in~\eqref{eq:extremal:m} and~\eqref{eq:extremal:cov}, respectively, is indeed feasible and optimal in~\eqref{eq:gelbrich:risk}.

It is instructive to contrast Theorem~\ref{thm:single:quad} with Theorem~\ref{theorem:convex:quadratic}, both of which provide exact tractable SDP reformulations for the problem of evaluating the worst-case risk of a quadratic loss function with respect to a type-2 Wasserstein ball. We highlight that the SDP~\eqref{eq:gelbrich:quad} derived in Theorem~\ref{thm:single:quad} for {\em elliptical} nominal distributions accommodates only two linear matrix inequalities, while the SDP~\eqref{eq:soroosh-convex} derived in Theorem~\ref{theorem:convex:quadratic} for {\em empirical} nominal distributions involves $N$ linear matrix inequalities and may thus be considerably harder to~solve.

\section{Performance Guarantees}
\label{sect:guarantees}
We now argue that for judiciously calibrated Wasserstein ambiguity sets, the worst-case risk~\eqref{eq:worst:risk} associated with a finite sample size $N$ provides an upper confidence bound on the true risk~\eqref{eq:risk} for all admissible loss functions (finite sample guarantee) and that the worst-case optimal risk~\eqref{eq:dro} converges almost surely to the true optimal risk~\eqref{eq:opt:risk} as $N$ tends to infinity (asymptotic guarantee). Intuitively, the finite sample guarantee ensures that the out-of-sample risk will fall short of the worst-case risk with high confidence when we implement an optimizer of the distributionally robust decision probelm~\eqref{eq:dro}, while the asymptotic guarantee formalizes the simple intuition that more data enables us to make better decisions.


Concentration inequalities for the nominal distribution $\Pnom_N$ and its moments can be used to derive finite sample and asymptotic guarantees. If $\Pnom_N$ is the empirical distribution, for instance, one can prove that~$\Pnom_N$ converges exponentially fast to the data-generating distribution $\PP$, in probability with respect to the Wasserstein distance, as $N$ tends to infinity.

\begin{theorem}[Concentration inequalities I]
    \label{theorem:concentration-empirical}
    Suppose that $\Pnom_N$ is the empirical distribution, while $p \neq m/2$, and the unknown true distribution $\PP$ is light-tailed in the sense that there exist $\alpha > p$ and $A>0$ such that $\EE^{\PP} [ \exp( \| \xi \|^\alpha) ] \leq A$. Then, there are constants $c_1, c_2>0$ that depend on $\PP$ only through $\alpha$, $A$, and $m$ such that for any $\eta \in (0, 1]$ the concentration inequality $\PP^N[\PP \in \mathbb{B}_{\eps, p}(\Pnom_N)] \geq 1-\eta$ holds whenever $\eps$ exceeds\footnote{Updated.}
    \be
    \label{eq:opt-radius-empirical}
    \eps_{N}(\eta) = \left\{ \begin{array}{ll} \DS \left(\frac{\log (c_1 /\eta)}{c_2N} \right)^{\min\{{1}/{m} ,{1}/{2}\}} & \DS \text{if } N \ge \frac{\log(c_1 /\eta)}{c_2}, \\[2ex]
    \DS \left(\frac{\log (c_1/ \eta) }{ c_2N} \right)^{{1}/{\alpha}} & \DS \text{if } N < \frac{\log(c_1 /\eta) }{c_2}.        \end{array}\right.
    \ee
\end{theorem}

Theorem~\ref{theorem:concentration-empirical} generalizes~\cite[Theorem~3.5]{esfahani2018data} to arbitrary $p\ge 1$ and is a direct consequence of~\cite[Theorem~2]{fournier2015rate}. The result remains valid for $p = m/2$ but with a more complicated formula for $\eps_{N}(\eta)$ \cite[Theorem~2]{fournier2015rate}. Intuitively, Theorem~\ref{theorem:concentration-empirical} asserts that any Wasserstein ball $\mathbb{B}_{\eps, p}(\Pnom_N)$ with radius~$\eps\ge \eps_{N}(\eta)$ represents a $(1-\eta)$-confidence set for the unknown data-generating distribution~$\PP$. For dimensions $m>2$, the critical radius~$\eps_{N}(\eta)$ of this confidence set decays as $\mathcal O(N^{-\frac{1}{m}})$. To reduce the critical radius by~$50\%$, the sample size must increase by  $2^m$. Unfortunately, this curse of dimensionality is fundamental, and the decay rate of $\eps_{N}(\eta)$ is essentially optimal; see \cite[\S~1.3]{fournier2015rate} or~\cite{weed2017sharp}.

The concentration inequality portrayed in Theorem~\ref{theorem:concentration-empirical} gives rise to the following finite sample guarantees~{\cite[Theorem~3.5]{esfahani2018data}}. 

\begin{theorem}[Finite sample guarantees I]
    \label{thm:finite-sample}
    Assume that all conditions of Theorem~\ref{theorem:concentration-empirical} hold and $\eps_{N}(\eta)$ is defined as in~\eqref{eq:opt-radius-empirical}. Then, for all $\eta \in (0, 1)$ and $\eps \geq \eps_N(\eta)$ we~have
    \begin{subequations}
    \begin{equation}
    \label{eq:generalization-1}
        \PP^N \Big\{  \risk(\PP, \ell) \leq \risk_{\eps, p}(\Pnom_N, \ell) \quad \forall \ell \in \mathcal L  \Big\} \geq 1 - \eta.
    \end{equation}
    Moreover, if $\ell\opt$ is an optimizer of the distributionally robust decision problem~\eqref{eq:dro}, which is a function of the training samples, then for all $\eta \in (0, 1)$ and $\eps \geq \eps_N(\eta)$ we have
    \begin{equation}
    \label{eq:generalization-2}
    \PP^N \Big\{\risk(\PP,\ell\opt) \leq \risk_{\eps,p}(\Pnom_N, \ell\opt) \Big\} \geq 1 - \eta.
    \end{equation}
    \end{subequations}
\end{theorem}

    Theorem~\eqref{thm:finite-sample} asserts that the worst-case risk~\eqref{eq:worst:risk} provides an upper confidence bound on the true risk~\eqref{eq:risk} under the unknown data-generating distribution uniformly across all loss functions~$\ell\in\Lc$. Moreover, it also asserts that the optimal value of the distributionally robust decision problem~\eqref{eq:dro} ({\em i.e.}, the worst-case optimal risk) provides an upper confidence bound on the out-of-sample performance of its optimizers. Note that the probabilities in~\eqref{eq:generalization-1} and~\eqref{eq:generalization-2} are evaluated under the distribution $\PP^N$ of the $N$ independent training samples.

\begin{remark}[Improved finite sample guarantees]
    \label{rem:improved guarantee}
    Requiring the Wasserstein ball to cover $\PP$ with high confidence is only a sufficient but not a necessary condition for the finite sample guarantees~\eqref{eq:generalization-1} and~\eqref{eq:generalization-2}. Indeed, these guarantees can be sustained even if the Wasserstein radius is reduced below $\eps_{N}(\eta)$, which is essentially the smallest radius for which the Wasserstein ball represents a $(1-\eta)$-confidence set for $\PP$. The minimal Wasserstein radius that preserves the finite sample guarantees~\eqref{eq:generalization-1} and~\eqref{eq:generalization-2} often decays significantly faster than $\mathcal O(N^{-\frac{1}{m}})$ without suffering from a curse of dimensionality. If $p=1$, the data-generating distribution is absolutely continuous with respect to the Lebesgue measure and the set $\Lc$ of admissible loss functions admits a smooth parameterization, for example, one can show that a Wasserstein radius of the order $\mathcal{O}(\sqrt{\log m/N})$ maintains finite sample guarantees akin to~\eqref{eq:generalization-1} and~\eqref{eq:generalization-2}, which is consistent with recent findings in the compressed sensing and high-dimensional statistics literature \cite[Theorem~1]{blanchet2016robust}. \hfill $\Box$
\end{remark}

As the number $N$ of training samples grows, one can simultaneously reduce the Wasserstein radius $\eps$ and the significance level $\eta$ without sacrificing the finite sample guarantees~\eqref{eq:generalization-1} and~\eqref{eq:generalization-2}, which allows us to prove asymptotic consistency \cite[Theorem~3.6]{esfahani2018data}. 

\begin{theorem}[Asymptotic consistency I]
    Assume that all conditions of Theorem~\ref{theorem:concentration-empirical} hold. Select $\eta_N \in (0, 1]$ and set $\eps_N= \eps_{N}(\eta_N)$ as in~\eqref{eq:opt-radius-empirical}, $N\in\mathbb N$, such that $\sum_{N=1}^\infty \eta_N < \infty$ and $\lim_{N \to \infty} \eps_N = 0$.\footnote{A possible choice is $\eta_N = \exp(-\sqrt{N})$.} If there exists $C > 0$ with $|\ell(\xi)| \leq C (1 + \| \xi \|^p)$ for all $\ell \in \mathcal L$ and $\xi \in \Xi$, then we have $\PP^\infty$-almost surely that $\risk_{\eps_N, p}(\Pnom_N, \mathcal L) \downarrow \risk(\PP, \mathcal L)$ as $N$ tends to infinity.
\end{theorem}

Next, we describe a concentration inequality for the sample mean and the sample covariance matrix that has ramifications for the Gelbrich risk minimization problem~\eqref{eq:gelbrich-dro}.

\begin{theorem}[Concentration inequalities II]
    \label{theorem:concentration-elliptical}
    Suppose that the unknown true distribution $\PP$ has mean vector $\mu$ and covariance matrix $\cov$ and that there are $\alpha > 2$ and $A>0$ such that $\EE^{\PP} [ \exp( \| \xi \|_2^\alpha) ] \leq A$. Then, there is $c>1$ that depends on $\PP$ only through $\mu$, $\cov$, $\alpha$, $A$, and $m$ such that for any $\eta \in (0, 1]$ the sample mean~$\msa$ and the sample covariance matrix~$\covsa$ satisfy the concentration inequality $\PP^N[(\mu,\cov)\in\mathcal U_\eps(\msa,\covsa)] \geq 1-\eta$ whenever $\eps$ exceeds
    \be
    \label{eq:opt-radius-elliptical}
    \eps_{N}(\eta) = \frac{\log (c /\eta)}{\sqrt{N}} .
    \ee
\end{theorem}

Theorem~\ref{theorem:concentration-elliptical} asserts that the uncertainty set $\mathcal U_\eps(\msa,\covsa)$ with radius $\eps\ge \eps_{N}(\eta)$ represents a $(1-\eta)$-confidence set for the mean vector and covariance matrix of the unknown data-generating distribution~$\PP$. The critical radius $\eps_{N}(\eta)$ of this confidence set decays as $\mathcal O(N^{-\frac{1}{2}})$ and is therefore---unlike the critical radius~\eqref{eq:opt-radius-empirical}---{\em not} subject to a curse of dimensionality. 

Theorem~\ref{theorem:concentration-elliptical} strengthens \cite[Theorem~2.3]{rippl2016limit}, which leverages a generalized central limit theorem to show that the type-2 Wasserstein distance between two normal distributions with true and empirical moments, respectively, decays {\em asymptotically} as $\mathcal O(N^{-\frac{1}{2}})$. A generalization of this result to elliptical distributions is discussed in \cite[Remark~2.4]{rippl2016limit}. 

\begin{theorem}[Finite sample guarantees II]
    \label{thm:finite-sample-Gelbrich}
    Assume that all conditions of Theorem~\ref{theorem:concentration-elliptical} hold and $\eps_{N}(\eta)$ is defined as in~\eqref{eq:opt-radius-elliptical}. Then, for all $\eta \in (0, 1)$ and $\eps \geq \eps_N(\eta)$ we~have
    \begin{subequations}
    \begin{equation}
    \label{eq:generalization-3}
        \PP^N \Big\{  \risk(\PP, \ell) \leq \overline \risk_{\eps}(\msa, \covsa, \ell) \quad \forall \ell \in \mathcal L  \Big\} \geq 1 - \eta.
    \end{equation}
    Moreover, if $\ell\opt$ is an optimizer of the Gelbrich risk optimization problem~\eqref{eq:gelbrich-dro}, which is a function of the training samples, then for all $\eta \in (0, 1)$ and $\eps \geq \eps_N(\eta)$ we have
    \begin{equation}
    \label{eq:generalization-4}
    \PP^N \Big\{\risk(\PP,\ell\opt) \leq \overline \risk_{\eps}(\msa, \covsa, \ell\opt) \Big\} \geq 1 - \eta.
    \end{equation}
    \end{subequations}
\end{theorem}

Theorem~\ref{thm:finite-sample-Gelbrich} asserts that the Gelbrich risk~\eqref{eq:gelbrich:risk} offers an upper confidence bound on the true risk~\eqref{eq:risk} under the unknown data-generating distribution uniformly across all loss functions. It also asserts that the optimal value of the Gelbrich risk optimization problem~\eqref{eq:gelbrich-dro} provides an upper confidence bound on the out-of-sample performance of its optimizers. 

 Asymptotic consistency of the Gelbrich risk optimization problem can only be established if the unknown true distribution is elliptical, $\Xi=\RR^m$ and all admissible loss functions are quadratic. By Theorem~\ref{thm:single:quad}, these conditions imply that the Gelbrich risk optimization problem~\eqref{eq:gelbrich-dro} is equivalent to the Wasserstein distributionally robust optimization problem~\eqref{eq:dro} equipped with a type-2 Wasserstein ball centered at an elliptical nominal distribution, where the Wasserstein distance is induced by the Euclidean norm.

\begin{theorem}[Asymptotic consistency II]
    Assume that all conditions of Theorem~\ref{theorem:concentration-elliptical} hold. Select $\eta_N \in (0, 1]$ and set $\eps_N= \eps_{N}(\eta_N)$ as in~\eqref{eq:opt-radius-elliptical}, $N\in\mathbb N$, such that $\sum_{N=1}^\infty \eta_N < \infty$ and $\lim_{N \to \infty} \eps_N = 0$. If $\PP$ is an elliptical distribution, $\Xi=\RR^m$ and $\ell(\xi)$ is quadratic for all $\ell\in\Lc$, then we have $\PP^\infty$-almost surely that $\overline \risk_{\eps_N, p}(\msa, \covsa, \mathcal L) \downarrow \risk(\PP, \mathcal L)$ as $N$ tends to infinity.
\end{theorem}

\section{Distributionally Robust Optimization in Machine Learning}
\label{sect:app}
We now demonstrate that the theory of data-driven distributionally robust optimization with Wasserstein ambiguity sets has interesting ramifications for statistical learning and motivates new approaches for addressing fundamental learning tasks such as classification (Section~\ref{sec:classification}), regression (Section~\ref{sec:regression}), maximum likelihood estimation (Section~\ref{sec:mle})  or minimum mean square error estimation (Section~\ref{sec:mmse}). We conclude with an overview of other applications of distributionally robust optimization in machine learning (Section~\ref{sec:other-applications}).

\subsection{Distributionally Robust Classification}
\label{sec:classification}
In binary classification problems the central object of study is a random vector~$\xi=(x,y)$, where~$x \in \RR^{n}$ is termed the {\em input}, and~$y\in\{-1,+1\}$ is referred to as the {\em output}. The distribution~$\PP$ of~$\xi$ is unknown but indirectly observable through finitely many training samples $\widehat\xi_i=(\widehat{x}_i, \widehat{y}_i)$, $i \in[N]$. The goal of binary classification is to predict the output~$y$ corresponding to a given input~$x$. The classifier with the lowest possible misclassification probability is the one that predicts $y=1$ if $\PP[y=1|x]\ge 0.5$ and $y=-1$ otherwise. Unfortunately, this classifier is not implementable when~$\PP$ is unknown.

Statistical learning aims to construct classifiers solely on the basis of the training data. One of the most popular approaches in practice is to construct a linear {\em scoring function}~$w^\top x$, encoded by a weight vector $w\in\RR^n$, and to predict $y$ as the sign of $w^\top x$. In hindsight, the prediction was correct if the actual output $y$ coincides with the predicted output $\mathop{\rm sign}(w^\top x)$ or, equivalently, if the product $y\cdot w^\top x$ is positive. The {\em realized} prediction error can thus be quantified by $L (y\cdot w^\top x)$, where $L(z)$ is some nonnegative and non-increasing univariate loss function that is large for negative and small for positive values of~$z$. Examples of popular loss functions are listed in Table~\ref{table:classification}. The best scoring function for a given choice of $L(z)$ is the one whose weight vector~$w$ minimizes the {\em expected} prediction error $\EE^\PP[L (y\cdot w^\top x)]$. Unfortunately, the expectation is evaluated under the unknown distribution $\PP$, and thus the optimal scoring function cannot be computed. Promising near-optimal scoring functions can be found, however, by solving a distributionally robust classification model that minimizes the worst-case expected prediction error with respect to a type-1 Wasserstein ball, that is, 
\begin{align}
\label{eq:classification} 
    \Inf_{w\in \RR^n} \Sup _ {\QQ \in \B_{\eps, 1}(\Pnom_N)} \EE ^ \QQ \left[ L (y\cdot w^\top x) \right],
\end{align}
where $\Pnom_N$ is the empirical distribution on the training samples. We assume here that all distributions in the Wasserstein ball are supported on $\Xi=\mathbb X \times\mathbb Y$, where $\mathbb X\subseteq\RR^n$ is convex and closed, while $\mathbb Y=\{-1,+1\}$. We also assume that the norm on the input-output space used in the definition of the Wasserstein distance is additively separable, that is, $\| \xi \| = \|x\|+ \frac{\kappa}{2} |y|$, where---by slight abuse of notation---$\|x\|$ stands for an arbitrary norm on the input space, while $\kappa>0$ quantifies the relative importance of outputs versus inputs.

\begin{table}[H]
    \centering
    \TABLE
    {Commonly used loss functions for binary classification. \label{table:classification}}
    {\begin{tabular*}{5in}{@{}l@{\qquad}l@{\qquad}l@{}}
            \hline
            \up name  & $L(z)$ & learning model \down\\
            \hline
            \up 
            hinge loss & $\max\{0, 1-z\}$ & support vector machine\\[1ex]
            smooth hinge  loss& $\left\{ 
            \begin{array}{ll}
            \frac{1}{2} - z & \text{if } z \leq 0 \\[0.5ex]
            \frac{1}{2} \left(1 - z\right)^2 & \text{if } 0 <z < 1 \\[0.5ex]
            0 & \text{if }z\ge 1
            \end{array} \right.$ & smooth support vector machine\\[4ex]
            logloss & $\log(1+ \exp(-z))$ & logistic regression
            \down\\
            \hline
    \end{tabular*}}
    {}
\end{table}

The classification model~\eqref{eq:classification} is easily recognized as an instance of the distributionally robust decision problem~\eqref{eq:dro} that optimizes over all (multivariate) loss functions of the form $\ell(\xi)=L(y\cdot w^\top x)$ parameterized by~$w\in\RR^n$. By leveraging Theorem~\ref{thm:type1}, problem~\eqref{eq:classification} can be recast as a finite convex program if $L(z)$ is convex and piecewise linear, while $\mathbb X$ is convex and closed. An alternative convex reformulation can be obtained from Theorem~\ref{thm:type1:convex} if $L(z)$ is convex (but not necessarily piecewise linear), while $\mathbb X=\RR^n$. For all univariate loss functions listed in Table~\ref{table:classification}, the convex reformulations of problem~\eqref{eq:classification} are equivalent to tractable conic programs. Explicit formulations of these conic programs are reported in \cite[\S~3.2]{shafieezadeh2017regularization}.

The distributionally robust classification problem~\eqref{eq:classification} encapsulates two interesting special cases. First, if the Wasserstein radius is set to $\eps=0$, then~\eqref{eq:classification} collapses to the standard empirical risk minimization problem that minimizes the average prediction error across the training samples. Moreover, if the parameter $\kappa$ appearing in the definition of the norm tends to infinity, then~\eqref{eq:classification} reduces to a classical {\em regularized} empirical risk minimization problem.

\begin{proposition}[Regularization by robustification]
\label{prop:regularization1}
    If $L(z)$ is any of the loss functions of Table~\ref{table:classification}, $\mathbb{X} = \RR^{n}$ and $\kappa = \infty$, then problem~\eqref{eq:classification} is equivalent to
\begin{align*}
    \inf_{w\in\RR^n}~ \frac{1}{N} \sum_{i =1}^N L(\widehat{y}_i \cdot w^\top \widehat{x}_i) + \eps\cdot \| w\|_*.
    \end{align*}
\end{proposition}

Recall that the norm $\| \xi \| = \|x\| + \frac{\kappa}{2} |y|$ on the input-output space encodes the transportation cost in the definition of the Wasserstein distance. Thus, $\kappa$ can be viewed as the cost of switching an output from $+1$ to $-1$ or vice versa. If $\kappa=\infty$, then all distributions in the Wasserstein ball are obtained by perturbing the empirical distribution along the input space because perturbations along the output space would be infinitely expensive. By setting $\kappa=\infty$, one thus postulates that there is only input uncertainty but no output uncertainty.

Proposition~\ref{prop:regularization1} gives commonly used regularization techniques a robustness interpretation, which applies under the premise that there is no output uncertainty. It identifies the regularization weight with the Wasserstein radius $\eps$ and the regularization function with the {\em dual} of the norm that determines the transportation cost along the input space. 

Proposition~\ref{prop:regularization1} can be deduced from Theorem~\ref{thm:type1} by observing that the Lipschitz modulus of the multivariate loss function $\ell(\xi)=L(y\cdot w^\top x)$ is given by $\Lip(L)\cdot\|w\|_*$ and that $\Lip(L)=1$ for all univariate loss functions of Table~\ref{table:classification}. Distributionally robust classification models with Wasserstein ambiguity sets were first studied in the context of logistic regression \cite{shafieezadeh2015distributionally}. Extensions to other classification models are discussed in \cite{blanchet2016robust, gao2017wasserstein, shafieezadeh2017regularization}.

\subsection{Distributionally Robust Regression}
\label{sec:regression}
The goal of regression is a to predict a {\em real} (as opposed to a categorical) output~$y\in\RR$ corresponding to a given input $x\in\RR^n$. The regressor that attains the lowest possible mean squared error is the one that predicts the output as $\EE^\PP[y|x]$. Unfortunately, this regressor is not implementable when the distribution~$\PP$ of the random vector $\xi=(x,y)$ is unknown.

In practice it is often convenient to construct a linear regressor that predicts the output by a linear function~$w^\top x$ encoded by a weight vector $w\in\RR^n$. The {\em realized} prediction error can thus be quantified by $L (w^\top x-y)$, where $L(z)$ is some nonnegative univariate loss function that is large when~$z$ deviates from~0. Examples of popular loss functions for regression are listed in Table~\ref{table:regression}. The best linear regressor that minimizes the {\em expected} prediction error $\EE^\PP[L (w^\top x-y)]$ cannot be computed when $\PP$ is unknown, but promising near-optimal linear regressors can be found by solving the distributionally robust regression model 
\begin{align} \label{eq:regression} 
\Inf_{w\in\RR^n} \Sup _ {\QQ \in \B_{\eps, p}(\Pnom_N)} \EE ^ \QQ \left[ L (w^\top x - y) \right],
\end{align}
which minimizes the worst-case expected prediction error in view of all distributions on a convex closed set $\Xi=\mathbb X \times\mathbb Y\subseteq \RR^n\times \RR$ within a type-$p$ Wasserstein ball around the empirical distribution on $N$ training samples. By Theorem~\ref{thm:type1}, problem~\eqref{eq:regression} can be reformulated as a finite convex program if $p=1$, $L(z)$ is convex and piecewise linear, and $\mathbb X$ and $\mathbb Y$ are convex and closed. A convex reformulation can also be obtained from Theorem~\ref{thm:type1} if $p=1$, $L(z)$ is convex (but not necessarily piecewise linear), while $\mathbb X=\RR^n$ and $\mathbb Y=\RR$. Moreover, problem~\eqref{eq:regression} can be reformulated as a finite convex program by using Theorem~\ref{theorem:convex:quadratic} if $p=2$, $L(z)$ is convex quadratic, $\mathbb X=\RR^n$ and $\mathbb Y=\RR$. For details see \cite[\S~3.1]{shafieezadeh2017regularization} and \cite[\S~3]{blanchet2016robust}.

\begin{table}[H]
    \centering
    \TABLE
    {Commonly used loss functions for regression \label{table:regression}}
    {\begin{tabular*}{5in}{@{}l@{\qquad}l@{\qquad}l@{\qquad}l@{}}
            \hline
            \up name & $L(z)$ & parameter & learning model \down\\
            \hline
            \up 
            squared error & $z^2$ & n/a & ordinary least squares \\[1ex]
            Huber loss & $\left\{ \begin{array}{ll} \frac{1}{2} \xi^2 & \text{if } |z| \leq \delta \\ \delta (|z| - \frac{1}{2} \delta) & \text{otherwise} \end{array} \right.$ & $\delta\in\RR_+$ & Huber regression\\[3ex]
            $\delta$-insensitive loss & $\max \{0, |z| - \delta\}$ & $\delta\in\RR_+$ & support vector regression\\[1ex]
            pinball loss & $\max\{-\delta z, (1-\delta)z\}$ & $\delta\in[0, 1]$ & quantile regression
            \down\\
            \hline
    \end{tabular*}}
    {}
\end{table}

Assume now that the norm on the input-output space satisfies $\| \xi \| = \|x\|+ \frac{\kappa}{2} |y|$, where $\|x\|$ is an arbitrary norm on the input space, while $\kappa>0$ quantifies the relative importance of outputs versus inputs. In the absence of output uncertainty (that is, for $\kappa=\infty$), there is again an intimate relation between robustification and regularization.

\begin{proposition}[Regularization by robustification]
\label{prop:regularization2}
    Assume that $\mathbb{X} = \RR^{n}$ and $\kappa = \infty$. If $L(z)$ is convex and Lipschitz continuous and $p=1$, then problem~\eqref{eq:regression} is equivalent to
    \begin{align*}
    \inf_{w\in\RR^n}~ \frac{1}{N} \sum_{i =1}^N L(w^\top \widehat{x}_i-\widehat{y}_i ) + \eps\cdot \Lip(L)\cdot \| w\|_*.
    \end{align*}
    Moreover, if $L(z)$ is the square error and $p=2$, then problem~\eqref{eq:regression} reduces to
    \begin{align*}
    \inf_{w\in\RR^n}~ \Big[ \Big(\frac{1}{N} \sum_{i =1}^N L(w^\top \widehat{x}_i-\widehat{y}_i )\Big)^{\frac{1}{2}}+ \eps \cdot \| w\|_*\Big]^2.
    \end{align*}
\end{proposition}

Proposition~\ref{prop:regularization2} asserts that if there is no output uncertainty, then the distributionally robust regression problem~\eqref{eq:regression} reduces to a regularized empirical risk minimization problem, where the regularization function is given by the dual of the norm on the input space. For Lipschitz continuous univariate loss functions $L(z)$ and for $p=1$, one simply minimizes the sum of the empirical risk and the regularization term weighted by the product of Wasserstein radius and the Lipschitz modulus of $L(z)$. Note that the Huber loss, the $\delta$-insensitive loss and the pinball loss are all Lipschitz continuous with Lipschitz moduli $\delta$, 1 and $\max\{\delta,1-\delta\}$, respectively. For the squared loss we need to set $p=2$ because the type-2 Wasserstein ball is the largest Wasserstein ball for which the worst-case expected loss is finite. In this case, one minimizes a combination of the square root of the empirical loss and the regularization term. If one measures distances in the input space using the $\infty$-norm, then this convex program reduces to the so-called generalized LASSO (Least Absolute Shrinkage and Selection Operator) estimation problem. For further details on distributionally robust regression see \cite{blanchet2016robust, gao2017wasserstein, shafieezadeh2017regularization}. 


\subsection{Distributionally Robust Maximum Likelihood Estimation}
\label{sec:mle}
Consider now the problem of estimating the mean vector $\mu\in\RR^m$ and the covariance matrix $\cov\in\mathbb S_+^m$ of a random vector $\xi\in\RR^m$ from independent training samples $\widehat \xi_i$, $i\in[N]$. The simplest estimators for $\mu$ and $\cov$ are the {\em sample mean} $\msa$ and the {\em sample covariance matrix} $\covsa$, which we define as the actual mean and covariance matrix of the empirical distribution,~{\em i.e.},
\[
\msa = \frac{1}{N} \sum_{i=1}^N \widehat{\xi}_i \qquad \text{and} \qquad  \covsa = \frac{1}{N} \sum_{i=1}^N (\widehat{\xi}_i - \msa)(\widehat{\xi}_i - \msa)^\top.
\]
While $\cov$ serves as an input for many problems in engineering, science or economics, it is often the precision matrix $\cov^{-1}$ that appears in their solutions. For example, in mean-variance portfolio analysis the portfolio variance to be minimized depends on the covariance matrix of the asset returns, while the optimal portfolio weights depend on the precision matrix. Similarly, linear discriminant analysis uses the covariance matrix of the features as an input and outputs a maximum likelihood classifier that depends on the precision matrix. Moreover, the optimal fingerprint method for climate change detection requires the covariance matrix of the internal climate variability as an input and outputs a climate change signal depending on the precision matrix. Thus, it is often more important to know the precision matrix than the covariance matrix. To ensure that the precision matrix is well defined, we will henceforth assume that $\cov\succ 0$. Unfortunately, the sample covariance matrix is rank-deficient in the big-data regime when the dimension of $\xi$ exceeds the sample size ($m>N$) even if $\cov$ has full rank. In this case, one cannot invert $\covsa$ to obtain a meaningful precision matrix estimator.

From now on we will assume that the unknown true distribution $\PP$ of $\xi$ is normal. Thus, the problem of maximizing the log-likelihood of the training samples reduces to the following convex program over all candidate mean vectors $\mu$ and precision matrices $X$ \cite[\S~7.1]{boyd2004convex}.
\be
\label{eq:unwise}
\Inf_{\mu \in \RR^m, \, X \in \mathbb S_{+}^m} \left\{ -\log\det X + \frac{1}{N} \sum_{i=1}^N (\widehat{\xi}_i - \mu)^\top X (\widehat{\xi}_i - \mu) \right\}
\ee
Unfortunately, this maximum likelihood estimation (MLE) problem is unbounded for $N\le m$ and (almost surely) solved by $\mu\opt=\msa$ and $X\opt=\covsa^{-1}$ for $N>m$. Thus, we fail again to find an estimator in the big-data regime and simply recover the sample mean and the sample covariance matrix in the small-data regime. To overcome this deficiency, we robustify the MLE problem against all distributions within a type-2 Wasserstein ball centered at the {\em normal} nominal distribution $\Pnom_N=\mathcal N(\msa,\covsa)$, that is, we solve the robust MLE problem
\be
\label{eq:wise}    
\begin{array}{cl}
    \Inf_{\mu \in\RR^m,\,X\in \mathbb{S}_{+}^m } \left\{ -\log \det X + \Sup_{\mathbb Q \in \B_{\eps, 2}(\Pnom_N)} \EE^{\mathbb Q} \left[(\xi-\mu)^\top X (\xi-\mu) \right] \right\}.
\end{array}
\ee
If $\eps=0$, then the robust MLE problem~\eqref{eq:wise} reduces to the nominal MLE problem~\eqref{eq:unwise} because---by the definition of the sample mean and the sample covariance matrix---the (normal) nominal distribution has the same first- and second-order moments as the (discrete) empirical distribution and because the loss function in the expectation is quadratic in $\xi$. One can show via Theorem~\ref{thm:single:quad} that~\eqref{eq:wise} is equivalent to a convex SDP with a determinant term in the objective function. Provided that the Wasserstein radius $\eps$ is strictly positive, this SDP is solvable even in the big-data regime when $m>N$. Thus, it yields a valid precision matrix estimator even if the sample covariance matrix is rank-deficient. Moreover, as SDPs are tractable, the optimal estimator can be computed in polynomial time. In fact, the SDP at hand is highly symmetric and can therefore even be solved in closed form \cite[Theorem~3.1]{nguyen2018distributionally}. 

\begin{theorem}[Wasserstein shrinkage estimator]
    \label{thm:wise:solution}
    If $\eps > 0$ and $\covsa\in\PSD^m$ admits the spectral decomposition $\covsa = \sum_{i=1}^m \eig_i\cdot v_i v_i^\top$ with eigenvalues $\eig_i \geq 0$ and corresponding orthonormal eigenvectors~$v_i$, $i \in[m]$, then the unique minimizer of the robust MLE problem~\eqref{eq:wise} is given by $\mu\opt = \msa$ and $X\opt = \sum_{i=1}^m x\opt_i \cdot v_i v_i^\top$, where 
    \begin{subequations}
        \label{eq:wise:analytical}
        \be
        \label{eq:wise:x}
        x\opt_i = \dualvar\opt \left[ 1 - \frac{1}{2} \left( \sqrt{\eig_i^2 (\dualvar\opt)^2  + 4 \eig_i \dualvar\opt } - \eig_i \dualvar\opt \right) \right] \qquad \forall i \in [m],
        \ee
        and $\dualvar\opt>0$ is the unique positive solution of the algebraic equation
        \be
        \label{eq:wise:gamma}
        \bigg( \eps^2 - \frac{1}{2} \sum_{i=1}^m \eig_i \bigg)  \dualvar - m + \frac{1}{2}  \sum_{i=1}^m   \sqrt{\eig_i^2 \dualvar^2  + 4 \eig_i \dualvar } = 0. 
        \ee
    \end{subequations}
\end{theorem}
Theorem~\ref{thm:wise:solution} asserts that the robust MLE estimator~$\mu\opt$ for the mean vector coincides with the sample mean~$\msa$. More interestingly, it further asserts that the robust MLE estimator~$X\opt$ for the precision matrix has the same eigenvectors~$v_i$ as the sample covariance matrix~$\covsa$, while its eigenvalues~$x\opt_i$ are obtained by applying the nonlinear transformation~\eqref{eq:wise:x} to the corresponding eigenvalues~$\eig_i$ of $\covsa$. This transformation involves a single unknown parameter~$\dualvar\opt$, which is the unique positive solution of the algebraic equation~\eqref{eq:wise:gamma}. As $X\opt$ is obtained by transforming the eigenvalues of the sample covariance matrix, it can be interpreted as a nonlinear shrinkage estimator. We thus refer to it as the {\em Wasserstein shrinkage estimator}.

As $X\opt$ and $\covsa$ share the same eigenvectors, $X\opt$ is {\em rotation-equivariant}, that is, the estimator applied to the rotated data $R\,\widehat\xi_i$, $i\in[N]$, coincides with the rotated estimator $RX\opt R$ of the original data for every possible rotation matrix $R$. Moreover, as all eigenvalues of $X\opt$ are strictly positive, the estimator is always invertible. Finally, Theorem~\ref{thm:wise:solution} indicates that~$X\opt$ can be computed highly efficiently by computing the spectral decomposition of $\covsa$ and by solving the scalar algebraic equation~\eqref{eq:wise:gamma}, which can be accomplished by bisection. 

One can show that the Wasserstein shrinkage estimator displays numerous desirable properties \cite[Proposition~3.5]{nguyen2018distributionally}. First, its eigenvalues $x_i\opt$ decrease with~$\eps$ and eventually converge to~0. This makes intuitive sense as for large values of $\eps$ nothing is known about~$\xi$, and thus the safest bet is that all of its components have high variance and low precision. Moreover, one can show that the order of the eigenvalues $x_i\opt$ matches the order of the inverse sample eigenvalues $1/\eig_i$ irrespective of $\eps>0$, which is expected in the absence of any structural information. Finally, one can show that the condition number of $X\opt$ decreases monotonically to~1 as $\eps$ grows. Thus, the condition number of $X\opt$ improves with the level of ambiguity.

A statistical theory that shows how to optimally choose $\eps$ is developed in \cite{blanchet2019inverse}. Surprisingly, the Wasserstein radius that attains the lowest possible out-of-sample loss scales as $\eps\propto 1/N$ instead of the canonical inverse square-root scaling, which may be expected for this problem. 

So far we have assumed that there is no structural information about the distribution of~$\xi$ besides normality. In some practical situation, however, the precision matrix $X$ may have a known sparsity pattern. Indeed, one can show that an element $X_{ij}$ of the precision matrix vanishes if and only if the random variables $\xi_i$ and $\xi_j$ are conditionally independent given all other components of $\xi$. Conditional independencies of this type naturally arise, for example, in the analysis of spatio-temporal data. In the presence of sparsity information, the robust MLE problem is still equivalent to a tractable SDP. Even though it loses its analytical solvability, one can devise a tailored sequential quadratic approximation algorithm with rigorous convergence guarantees to solve the problem numerically, see~\cite[\S~4]{nguyen2018distributionally}.

\subsection{Distributionally Robust Minimum Mean Square Error Estimation}
\label{sec:mmse}

Consider next the problem of estimating a signal $x\in \RR^{m_x}$ from a noisy observation $y \in\RR^{m_y}$ under the premise that the distribution of the random vector $\xi = [x^\top, y^\top]^\top \in \RR^m$, $m=m_x+m_y$, is ambiguous and only known to belong to a type-2 Wasserstein ball centered at an elliptical nominal distribution $\Pnom_N =\mathcal E_g(\msa, \covsa)$ with nominal mean vector $\msa \in\RR^m$, nominal covariance matrix $\covsa\in\PSD^m$ and density generator $g$. This elementary problem is fundamental for numerous applications in engineering ({\em e.g.}, linear systems theory \cite{golnaraghi2017automatic, ogata2009modern}), econometrics ({\em e.g.}, linear regression \cite{stock2015introduction, wooldridge2010econometric}, time series analysis \cite{chatfield2016analysis, hamilton1994time}), machine learning and signal processing ({\em e.g.}, Kalman filtering \cite{kay1993fundamentals, murphy2012machine, oppenheim2015signals}) or information theory ({\em e.g.}, multiple-input multiple-output systems \cite{cover2012elements, mackay2003information}), etc. To formalize the estimation problem, we define an estimator as a measurable function $\psi(y)$ that maps the observation $y$ to a prediction of the signal $x$, and we denote by $\Psi$ the family of all possible estimators. Moreover, we define the distributionally robust {\em minimum mean square error} (MMSE) estimator as an optimizer of 
\begin{equation}
\label{eq:mmse}
\Inf_{\psi\in\Psi} \Sup_{\mathbb Q \in \B_{\eps, 2}(\Pnom_N)} \EE^{\mathbb Q} \left[ \| x - \psi(y) \|_2^2 \right].
\end{equation}
Note that~\eqref{eq:mmse} constitutes an infinite-dimensional functional optimization problem and thus appears to be hard. However, by establishing a minimax theorem for~\eqref{eq:mmse} and exploiting the properties of elliptical distributions, one can show that the outer infimum in~\eqref{eq:mmse} is attained by an {\em affine} estimator. Combining this structural insight with Theorem~\ref{thm:single:quad} allows us to prove that the estimation problem~\eqref{eq:mmse} is in fact equivalent to a convex program~\cite{shafieezadeh2018wasserstein}.

\begin{theorem}[Distributionally robust MMSE estimator]
If $\covsa\succ 0$, then the estimation problem~\eqref{eq:mmse} is equivalent to the nonlinear convex SDP
\begin{equation}
\label{eq:mmse:dual}
\begin{split}
\max \quad & f(S) = \Tr{S_{xx} - S_{xy} S_{yy}^{-1} S_{yx} } \\
\st \quad & S= \begin{bmatrix} S_{xx} & S_{xy} \\ S_{yx} & S_{yy} \end{bmatrix} \in \PSD^m, \quad S_{xx} \in \PSD^{m_x}, \quad S_{yy} \in \PSD^{m_y}, \quad S_{xy}=S_{yx}^\top \in \RR^{m_x \times m_y} \\
& \Tr{S + \covsa - 2 \left( \covsa^{\frac{1}{2}} S \covsa^{\frac{1}{2}} \right)^{\frac{1}{2}}} \leq \eps^2,\quad S  \succeq \eig_{\min}(\covsa) I,
\end{split}
\end{equation}
which is always solvable. If $S\opt$, $S_{xx}\opt$, $S_{yy}\opt$ and $S_{xy}\opt$ are optimal in~\eqref{eq:mmse:dual}, while $\msa_x\in\RR^{m_x}$ and $\msa_y\in\RR^{m_y}$ are the (known) mean vectors of $x$ and $y$ under $\Pnom_N$, respectively, then the affine function $\psi\opt(y) = S_{xy}\opt (S_{yy}\opt)^{-1} (y - \msa_y) + \msa_x$ is a distributionally robust MMSE estimator.
\end{theorem}

It is possible to eliminate all nonlinearities in~\eqref{eq:mmse:dual} by using Schur complements and to reformulate the nonlinear convex SDP as a standard linear SDP, which is formally tractable. However, larger problem instances quickly exceed the capabilities of general-purpose solvers. Instead, there is merit in addressing the nonlinear SDP~\eqref{eq:mmse:dual} directly with a customized first-order Frank-Wolfe algorithm, which starts at $S^{(0)}=\covsa$ and constructs iterates
\[
    S^{(k+1)}=\alpha_k D^{(k)} + (1-\alpha_k)S^{(k)} \qquad \forall k=0,1,2,\ldots
\]
with stepsize $\alpha_k$, where $D^{(k)}\in\mathbb S^m$ is the unique solution of the direction-finding subproblem
\be \label{eq:mmse:direction}
\begin{split}
    \max_{D\in\mathbb S^m} \quad & \Tr{D \; \nabla f(S^{(k)})} \\
    \st \quad & \Tr{D + \covsa - 2 \left( \covsa^{\frac{1}{2}} D \covsa^{\frac{1}{2}} \right)^{\frac{1}{2}}} \leq \eps^2,\quad D  \succeq \eig_{\min}(\covsa) I .
\end{split}
\ee
The Frank-Wolfe algorithm is highly efficient because the direction-finding subproblem~\eqref{eq:mmse:direction}, which linearizes the objective function $f(S)$ of~\eqref{eq:mmse:dual} around the current iterate $S^{(k)}$, can be solved in closed form. Indeed, using Theorem~\ref{thm:extremal:easy} one can show that~\eqref{eq:mmse:direction} is solved by
\[
    D^{(k)} = (\gamma\opt)^2 \left( \gamma\opt I- \nabla f(S^{(k)}) \right)^{-1} \covsa \left( \gamma\opt I- \nabla f(S^{(k)}) \right)^{-1},
\]
where $\gamma\opt$ is the unique solution with $\gamma\opt I \succ \nabla f(S^{(k)})$ of the algebraic equation
\[
    \Tr{ \covsa \left(I- \gamma (\gamma I -  \nabla f(S^{(k)}))^{-1}\right)^2}=\eps^2,
\]
which can be solved via bisection \cite[Theorem~3.2]{shafieezadeh2018wasserstein}. For a judiciously chosen step-size rule, the Frank-Wolfe algorithm also offers rigorous convergence guarantees \cite[Theorem~3.3]{shafieezadeh2018wasserstein}.

\begin{theorem}[Convergence analysis]
If $\covsa\succ 0$, $\eps>0$ and $\alpha_k=\frac{2}{k+2}$ for every $k\in\mathbb N$, then the $k^{\rm th}$ iterate $S^{(k)}$ of the Frank-Wolfe algorithm is feasible in~\eqref{eq:mmse:dual} and satisfies $f(S\opt) - f(S^{(k)}) \le  \frac{C}{k+2}$, where $C$ depends only on $\covsa$ and $\eps$, and $S\opt$ is a maximizer of~\eqref{eq:mmse:dual}.
\end{theorem}

In some applications one has additional structural information about the relation between the signal~$x$ and the observation~$y$ ({\em e.g.}, the measurement noise may be known to be independent of the signal, or the observation may be governed by a linear measurement model, etc.). Such structural information can be used to restrict the Wasserstein ambiguity set in~\eqref{eq:mmse}, thereby  reducing the conservativeness of the distributionally robust MMSE estimator~\cite{nguyen2019bridging}.

\subsection{Other Applications in Machine Learning}
\label{sec:other-applications}
Ideas from distributionally robust optimization also permeate several other areas of statistics and machine learning. For example, a distributionally robust optimization model involving two Wasserstein balls centered at two distinct empirical distributions can be used to develop a computationally tractable convex approximation for the {\em minimax robust hypothesis testing problem} that aims to minimize the maximum of the worst-case type-I and type-II errors of a prescribed hypothesis test \cite{gao2018robust}. Another example is {\em data-driven inverse optimization}, where one  observes random signals as well as optimal solutions of an optimization problem parameterized by these signals. The aim is to predict the solution corresponding to a new unseen signal from $N$ independent historical observations without any knowledge of the optimization problem's objective function. This problem can be framed as a {\em structural regression problem} that minimizes the worst-case expected prediction loss with respect to a Wasserstein ambiguity set over a space of candidate objective functions~\cite{esfahani2018inverse}. Data-driven inverse optimization lends itself, for example, to learning the purchasing behavior of consumers, the production costs of electricity generators, the route choice preferences of passengers in a multimodal transportation system or the hidden optimality principles governing a biological system. As a third example, distributionally robust optimization models with Wasserstein ambiguity sets can be used to efficiently compute the {\em worst-case misclassification probability} of a given classifier, which amounts to evaluating the worst-case expectation of the (nonconvex) zero-one loss \cite{shafieezadeh2017regularization, shafieezadeh2015distributionally}. Using similar techniques, one can also efficiently compute the worst-case probability of an undesirable event described by the conjunction or disjunction of several linear inequalities for the random vector~$\xi$ \cite{hanasusanto2015perspective, esfahani2018data}. If the undesirable event can be influenced so as drive its worst-case probability below a prescribed tolerance, we face a {\em distributionally robust chance constraint}. Even though distributionally robust chance constrained programs with Wasserstein ambiguity sets around the empirical distribution are intractable in general, they are sometimes equivalent to mixed-integer linear programs that can be solved with off-the-shelf software \cite{chen2018drocc, xie2018drocc}. In contrast, distributionally robust chance constrained programs with moment ambiguity sets can often be reformulated as (or tightly approximated by) tractable conic programs \cite{calafiore2006distributionally, cheung2012linear, hanasusanto2015perspective, zymler2013distributionally}.

To conclude, we highlight two opportunities for tailoring a distributionally robust decision problem with a Wasserstein ambiguity set around the empirical distribution 
to a given training dataset. Recall first that finite sample guarantees hold whenever~$\eps$ is large enough for the Wasserstein ball to contain the unknown data-generating distribution with high confidence~$1-\beta$. Recall also that the distributionally robust decision problem can often be reformulated as a tractable convex program whose size scales with the sample size $N$. If the computational burden is unmanageable for the given sample size, we can select $K\ll N$, approximate $\Pnom_N$ with the closest $K$-point distribution $\QQ\opt_K$ in Wasserstein distance and replace the original Wasserstein ball of radius $\eps$ around $\Pnom_N$ with a new inflated Wasserstein ball of radius $\eps + W_p(\Pnom_N, \QQ\opt_K)$ around $\QQ\opt_K$. By construction, the inflated Wasserstein ball contains the data-generating distribution with the same confidence $1-\beta$. But the size of the corresponding decision problem is only proportional to $K$. This approach provides a systematic method for reducing the computational burden without sacrificing robustness guarantees (but at the expense of increasing the model's level of conservatism). The approximation of a rich $N$-point distribution with a sparse $K$-point distribution is referred to as {\em scenario reduction} in the stochastic programming literature. While the exact computation of~$\QQ\opt_K$ is hard, there exist efficient approximation algorithms for scenario reduction~\cite{rujeerapaiboon2018scenario}.

An important input for any distributionally robust optimization model with a Wasserstein ambiguity set is the norm that determines the transportation cost in the definition of the Wasserstein distance. The flexibility to choose this norm could be exploited to improve the out-of-sample performance of the model's optimizers. A method for learning the best Mahalanobis norm from the training data is described in~\cite{blanchet2017data}. It is shown that this metric learning framework encompasses {\em adaptive regularization} as a special case.

\medspace 

\emph{Acknowledgments.} This research was funded by the SNSF grant BSCGI0\_157733. We are grateful to Erick Delage, Bart Van Parys and Shuhao Yan for pointing out errors in the published version \cite{kuhn2019wasserstein} of this paper.

\begin{APPENDIX}{}
    
    \section{Elliptical Distributions}
    \label{sect:elliptical}
    
    We say that $\QQ=\mathcal{E}_g(\mu, \cov)$ is an {\em elliptical} probability distribution if it has a density function of the form $f(\xi) = C \det(\cov)^{-1} \, g((\xi-\mu) \cov^{-1} (\xi-\mu))$ with density generator $g(u)\ge 0$ for all $u\ge 0$, normalization constant $C>0$, mean vector $\mu\in\RR^m$ and covariance matrix $\cov\in\mathbb{S}_{++}^m$.
    
    \begin{table}[H]
        \TABLE
        {Examples of elliptical distributions\label{tabel:elliptical}.}
        {\begin{tabular*}{5in}{@{}l@{\qquad}l@{\qquad}l@{}}
                \hline
                \up
                distribution family & density generator $g(u)$ & normalization constant $C$ \down \\ \hline
                \up Gaussian distribution& $\exp(-u/2)$ & $(2\pi)^{-m/2}$\\[1.5ex]
                Logistic distribution & $\DS  \frac{\exp(-u)}{(1+\exp(-u))^2}$ & $ \DS \frac{\pi^{m/2}}{\Gamma(m/2)} \int_0^\infty \frac{y^{m/2 - 1} \exp(-y)}{(1 + \exp(-y))^2} \mathrm{d}y$   \\[2ex]
                $t$-distribution & $\DS  \left( 1 + u / (\nu-2) \right)^{-\frac{m + \nu}{2}}$ & $\DS \frac{\nu^{m/2} \Gamma((m+\nu)/2)}{\pi^{m/2} \Gamma(\nu/2) (\nu-2)^m} $ \down\\[2ex] \hline
        \end{tabular*}}
        {$\nu > 2$ denotes the degrees of freedom of the $t$-distribution, and $\Gamma$ is the gamma~function.}
    \end{table}    
    
    \section{Conjugates, Support Functions and Dual Norms}
    \label{sect:functions}
    
    The conjugate of an extended real-valued function $\ell(\xi)$ on $\RR^m$ is a function $\ell^*(z)$ on $\RR^m$ defined through $\ell^*(z) = \sup_{\xi} z^\top \xi - \ell(\xi)$. If $\ell(\xi)$ is proper, convex and lower semicontinuous, then the conjugate of the conjugate coincides with the initial function, that is, $\ell^{**}(\xi)=\ell(\xi)$ \cite[\S~12]{rockafellar1997convex}.

    \begin{table}[H]
        \centering
        \TABLE
        {Examples of conjugates. \label{table:conjugate}}
        {\begin{tabular*}{5in}{@{}l@{\qquad}l@{\qquad}l@{\qquad}l@{}}
                \hline
                \up $\ell(\xi)$ & $\mathrm{dom}(\ell)$ & $\ell^*(z)$ & $\mathrm{dom}{(\ell^*)}$ \down\\
                \hline
                \up $ a^\top \xi + b$ & $\RR^m$ &  $ \delta_{\{a\}}(z) - b$ & $\{a\}$ \\
                $ \frac{1}{2} \xi^\top A \xi + a^\top \xi + b$ & $\RR^m$ & $ \frac{1}{2} (z- a)^\top A^\dagger (z-a) - b$ & $ \{a\} + \mathrm{range(A)}$\\
                $\log(1+\exp(-\xi))$ & $\RR$ & $z \log z + (1-z) \log (1-z)$ & $[0, 1]$\\
                $\exp(\xi)$ & $\RR$ & $z \log z - z$ & $\RR_+$\down\\
                $ \frac{1}{p}\| \xi \|^p $ & $\RR^m$ & $ \frac{1}{q}\| \xi \|_*^q$
                & $\RR^m$ \\
                $ \| \xi \| $ & $\RR^m$ & $\delta_{\mathcal B^*_1(0)}(z)$  
                & $\mathcal B^*_1(0)$ \\
                \hline
        \end{tabular*}}
        {Assume that $A \in \PSD^m$, $a \in \RR^m$, $b \in \RR$ and $p,q\ge 1$ with $\frac{1}{p}+\frac{1}{q}=1$. Moreover, denote by $\mathcal B^*_1(0)=\{z\in\RR^m: \| z\|_*\le 1\}$ the standard ball with respect to the dual norm on $\RR^m$.}
    \end{table}

    The indicator function of a set $\Xi\subseteq \RR^m$ is a function $\delta_\Xi(\xi)$ on $\RR^m$ defined through $\delta_\Xi(\xi)= 0$ if $\xi\in\Xi$ and $\delta_\Xi(\xi)=\infty$ if $\xi\notin \Xi$. The support function of $\Xi\subseteq \RR^m$ is a function $\sigma_\Xi(z)$ on $\RR^m$ defined through $\sigma_\Xi(z) = \sup_{\xi \in \Xi}  z^\top \xi$. The support function of $\Xi$ coincides with the conjugate of the indicator function of $\Xi$, that is, $\delta_\Xi^*(z)=\sigma_\Xi(z)$. If $\Xi$ is convex and closed, then the conjugate of the support function of $\Xi$ coincides with the indicator function of $\Xi$, that is, $\sigma_\Xi^*(\xi)=\delta_\Xi(\xi)$ \cite[\S~13]{rockafellar1997convex}.

    \begin{table}[H]
        \centering
        \TABLE
        {Examples of support functions\label{table:support}.}
        {\begin{tabular*}{5in}{@{}l@{\qquad}l@{\qquad}l@{}}
                \hline
                \up $\Xi$ & $\sigma_\Xi(z)$ & $\mathrm{dom}(\sigma_{\Xi})$\down\\
                \hline
                \up $\{ \xi : \, \| \xi \| \leq b \}$ & $ b \|z\|_*$ & $\RR^m$ \\
                $\{ \xi : \, C \xi \leq d \}$ & $\inf\{ \lambda^\top d : \lambda \in \RR_+^l, C^\top \lambda = z \}$ & $\{C^\top\lambda : \lambda \in \RR_+^l \}$ \\
                $\{ \xi : \, f(\xi) \leq 0 \}$ &  $\inf \{ \lambda f^*(z / \lambda): \lambda \in \RR_+^l \}$ & $- \mathop{\rm recc}(f)^*$ \\
                $\{ \xi : \, \xi \in \Xi_k ~ \forall k \in [K]\} $ & $\inf \{ \sum_{k=1}^K \sigma_{\Xi_k}(z_k): \sum_{k=1}^K z_k = z \}$ & $-\bigcap_{k\in[K]} \mathop{\rm recc}(\Xi_k)^*$ \down\\
                \hline
        \end{tabular*}}
        {Assume that $b \in \RR_+$, $C \in \RR^{l \times m}$ and $d \in \RR^l$. Let $f(\xi)$ be a closed, proper and convex function, and let $\Xi_k$, $k\in[K]$, be convex closed sets with nonempty intersection. Denote by $\mathop{\rm recc}(f)^*$ and $\mathop{\rm recc}(\Xi_k)^*$ the cones dual to the recession cones of the function $f(\xi)$ and the set $\Xi_k$, respectively.} 
    \end{table}

    If $ \| \cdot \| $ is a norm on~$\RR^m$, then its dual norm $\|\cdot\|_*$ on $\RR^m$ is defined through $ \| z \|_* =\sup_{\| \xi \| \leq 1} z^\top \xi$. The dual norm of the dual norm coincides with the original norm, that is, $\|\cdot\|_{**}=\|\cdot\|$ \cite[\S~15]{rockafellar1997convex}.
    
    \begin{table}[H]
        \centering
        \TABLE
        {Examples of dual norms\label{table:norm}.}
        {\begin{tabular*}{5in}{@{}l@{\qquad}l@{\qquad}l@{}}
                \hline
                \up $\| \xi \|$ & $\| z \|_*$ & comment\down\\
                \hline
                $ \| \xi \|_p$ & $ \| z \|_q$ & standard $p$-norms \\
                $\| \xi \|_1$ & $ \| z \|_\infty$ & limiting case when $p\downarrow 1$ and $q\uparrow\infty$ \\
                $ \alpha \| \xi \|_p $ & $ \frac{1}{\alpha}\| z \|_q $ & scaled $p$-norms \\
                $ \| A \xi \|_p $ & $ \|A^{-1} z \|_q $ & scaled $p$-norms \\
                $ \sum_{k \in [K]} \| \xi_k \|_{p_k} $ & $\max_{k \in [K]} \|z_k \|_{q_k}$ & additively separable norms\\
                \hline
        \end{tabular*}}
        {Assume that $p,q\ge 1$ with $\frac{1}{p}+\frac{1}{q}=1$, $\alpha>0$, $A\in\mathbb S_{++}^m$, and $p_k,q_k\ge 1$ with $\frac{1}{p_k}+\frac{1}{q_k}=1$ for all $k\in[K]$. Moreover, $\xi = (\xi_1, \ldots, \xi_K)$ and $z = (z_1, \ldots, z_K)$, where $\xi_k,z_k \in \RR^{m_k}$ and $\sum_{k=1}^K m_k = m $.}
    \end{table}
    
\end{APPENDIX}


\bibliographystyle{./TutORials}
\bibliography{bibliography}

\end{document}